\documentclass[journal]{IEEEtran}

\ifCLASSINFOpdf

\else

\fi

\UseRawInputEncoding
\usepackage{graphicx}

\usepackage{amssymb}
\usepackage{amsthm}
\usepackage{amsmath}
\usepackage{color}
\usepackage{setspace}
\usepackage{url}
\usepackage{makecell}
\usepackage{algorithm}
\usepackage{algorithmic}
\usepackage{graphicx}
\usepackage{subfigure}
\usepackage{multirow}
\usepackage{array}
\usepackage[font=small,labelfont=bf,tableposition=top]{caption}
\usepackage[numbers,sort&compress]{natbib}
\usepackage{booktabs}
\usepackage{float} 
\usepackage{threeparttable}
\usepackage{stfloats}
\usepackage{color}
\renewcommand{\baselinestretch}{1.0}
\usepackage{amsmath}
\usepackage{makecell}
\usepackage{setspace}
\usepackage[table]{xcolor}
\usepackage{hyperref}
\begin{document}

%\begin{spacing}{2.0}

\title{Deep Learning for Logo Detection: A Survey}

% \author{Jing Wang, Weiqing Min, \textit{Member, IEEE}, Sujuan Hou, \textit{Member, IEEE}, Shengnan Ma, \\ Yuanjie Zheng, \textit{Member, IEEE}, Shuqiang Jiang, \textit{Senior Member, IEEE}}
        % <-this % stops a space

%}
% The paper headers

\author{Sujuan Hou, \textit{Member, IEEE}, Jiacheng Li, Weiqing Min, \textit{Senior Member, IEEE}, Qiang Hou, \\ Yanna Zhao, \textit{Member, IEEE}, Yuanjie Zheng, \textit{Member, IEEE}, and Shuqiang Jiang, \textit{Senior Member, IEEE}\thanks{This work was supported by the National Nature Science Foundation of China (No.62072289, 61972378, U193620), CAAI-Huawei MindSpore Open Fund. S. Hou, J. Li, Q. Hou, Y. Zhao and Y. Zheng are School of Information Science and Engineering, Shandong Normal University, Shandong, 250358, China. Email: sujuanhou@sdnu.edu.cn, 2021317140@stu.sdnu.edu.cn, 2019309052@stu.sdnu.edu.cn, yannazhao@sdnu.edu.cn, and  zhengyuanjie@gmail.com. W. Min and S. Jiang are with the Key Laboratory of Intelligent Information Processing, Institute of Computing Technology, Chinese Academy of Sciences, Beijing, 100190, China, and also with University of Chinese Academy of Sciences, Beijing, 100049, China. Email: minweiqing@ict.ac.cn, and sqjiang@ict.ac.cn. }}

\maketitle

% \markboth{IEEE Transactions on Image Processing,~Vol.~X,
% No.~XX,~Month~Year}{}
% \providecommand{\keywords}[1]{\textbf{\textit{Index terms---}} #1}
% \author{Weiqing Min$^{1}$,  Bing-Kun Bao$^{1}$,  Changsheng Xu$^{1,2}$\\$^{1}$National Lab of Pattern Recognition, Institute of Automation, CAS, \\Beijing 100190, China\\$^{2}$China-Singapore Institute of Digital Media, Singapore, 139951, Singapore}
% \institute{$^{1}$National Lab of Pattern Recognition, Institute of Automation, CAS, \\Beijing 100190, China\\$^{2}$China-Singapore Institute of Digital Media, Singapore, 119613, Singapore}

% \{wqmin,csxu\}@nlpr.ia.ac.cn, bingkunbao@gmail.com

%  make the title area

\begin{abstract}
When logos are increasingly created, logo detection has gradually become a research hotspot across many domains and tasks. Recent advances in this area are dominated by deep learning-based solutions, where many datasets, learning strategies, network architectures, etc. have been employed. This paper reviews the advance in applying deep learning techniques to logo detection. Firstly, we discuss a comprehensive account of public datasets designed to facilitate performance evaluation of logo detection algorithms, which tend to be more diverse, more challenging, and more reflective of real life. Next, we perform an in-depth analysis of the existing logo detection strategies and the strengths and weaknesses of each learning strategy. Subsequently, we summarize the applications of logo detection in various fields, from intelligent transportation and brand monitoring to copyright and trademark compliance. Finally, we analyze the potential challenges and present the future directions for the development of logo detection to complete this survey. 

%Logo detection has gradually become a research hotspot in the field of computer vision and multimedia for its wide applications. With the rapid development of deep learning, logo detection has made considerable progress. In this paper, we comprehensively review the state-of-the-art methods in logo detection. We first provide a comprehensive overview of available datasets designed to facilitate performance evaluation of logo detection algorithms, which tend to be more diverse, more challenging, and more reflective of real life. We then review the state-of-the-art in the field of logo detection and discuss their contributions. We also highlight the major limitations of existing studies in addressing the challenges in this field. Moreover, we summarize the applications of logo detection in various fields, from intelligent transportation, brand monitoring to copyright and trademark compliance. Finally, we propose the future directions for the development of logo detection. This survey aims to encourage researchers to propose more efficient algorithms for logo detection.

\end{abstract}


% Note that keywords are not normally used for peerreview papers.
%\begin{IEEEkeywords}
%IEEEtran, journal, \LaTeX, paper, template.
%\end{IEEEkeywords}

\IEEEpeerreviewmaketitle

\section{Introduction}

Logos usually consist of texts, shapes, images, or their combination. Logo detection benefits a wide range of applications in different areas, such as intelligent transportation~\cite{26Yu2021,surwase2021multi}, social media monitoring~\cite{hu2020multimodal}, and infringement detection~\cite{patalappa2021robust,chen2021robust}. Meanwhile, some competitions have emerged, such as Robust Logo Detection Grand Challenge~\cite{chen2021robust,jia2021effective,leng2021gradient,xu2021simple} and Few-shot Logo Detection~\cite{Tong2022ICME2F}.

The main task of logo detection is to determine the location of a specific logo in images/videos and identify them. Although it may be regarded as a particular object task, logo detection in real-world images can be pretty challenging since numerous brands may have highly diverse contexts, varied scales, changes in illumination, size, resolution, and even non-rigid deformation (as shown in Fig. 1).

\begin{figure}[htbp!]  
	\centering
	\hspace{-3mm}
	\subfigure[dim light]{
	
		\includegraphics[width=2.8cm]{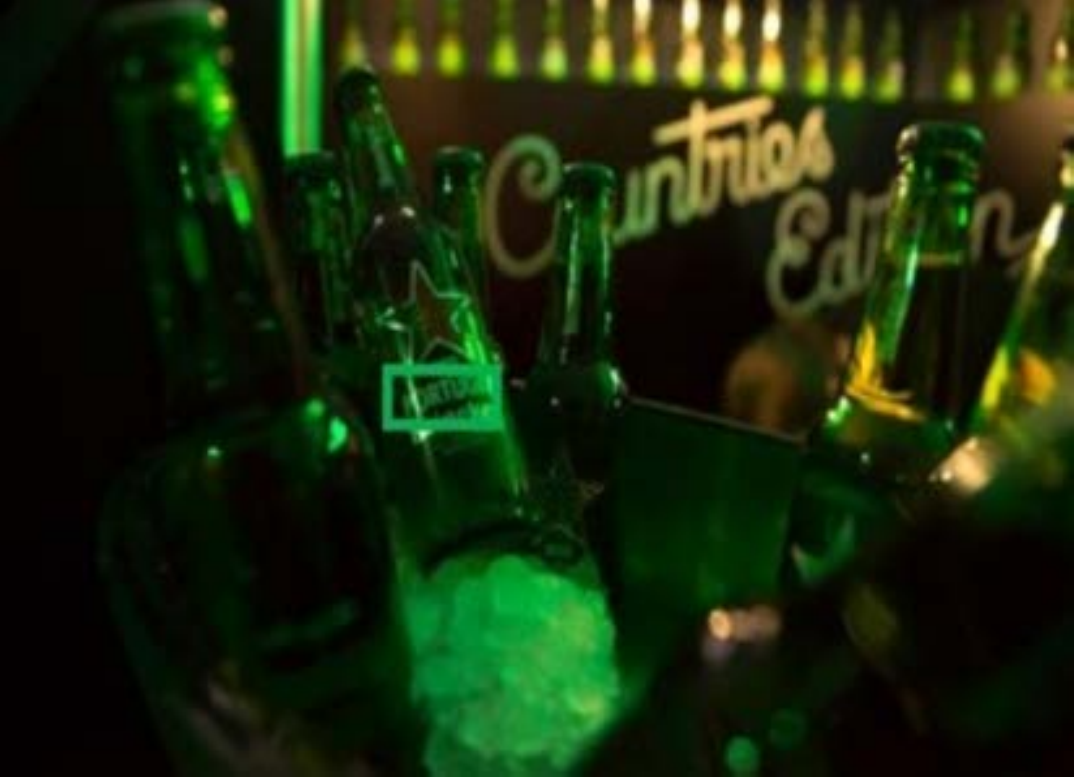}
		%\caption{fig1}
	}
	\hspace{-4mm}
	\subfigure[image rotation]{
	
		\includegraphics[width=2.8cm]{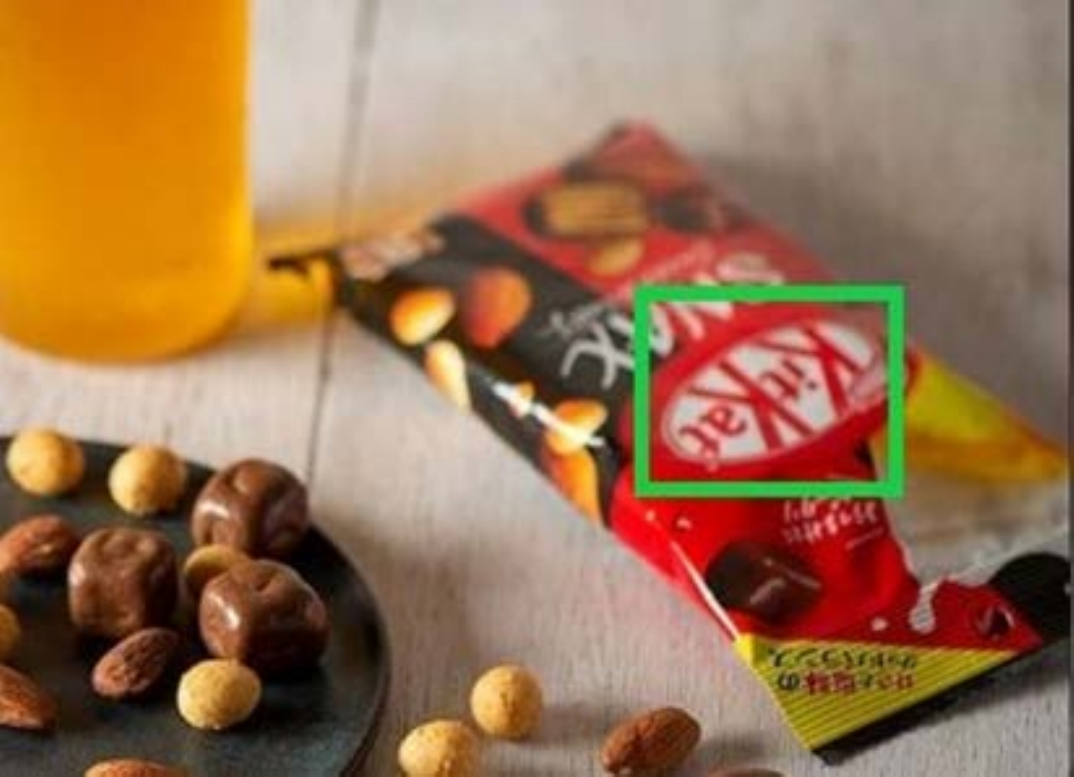}
	}
	\hspace{-4mm}
	\subfigure[small-scale logos]{

		\includegraphics[width=2.8cm]{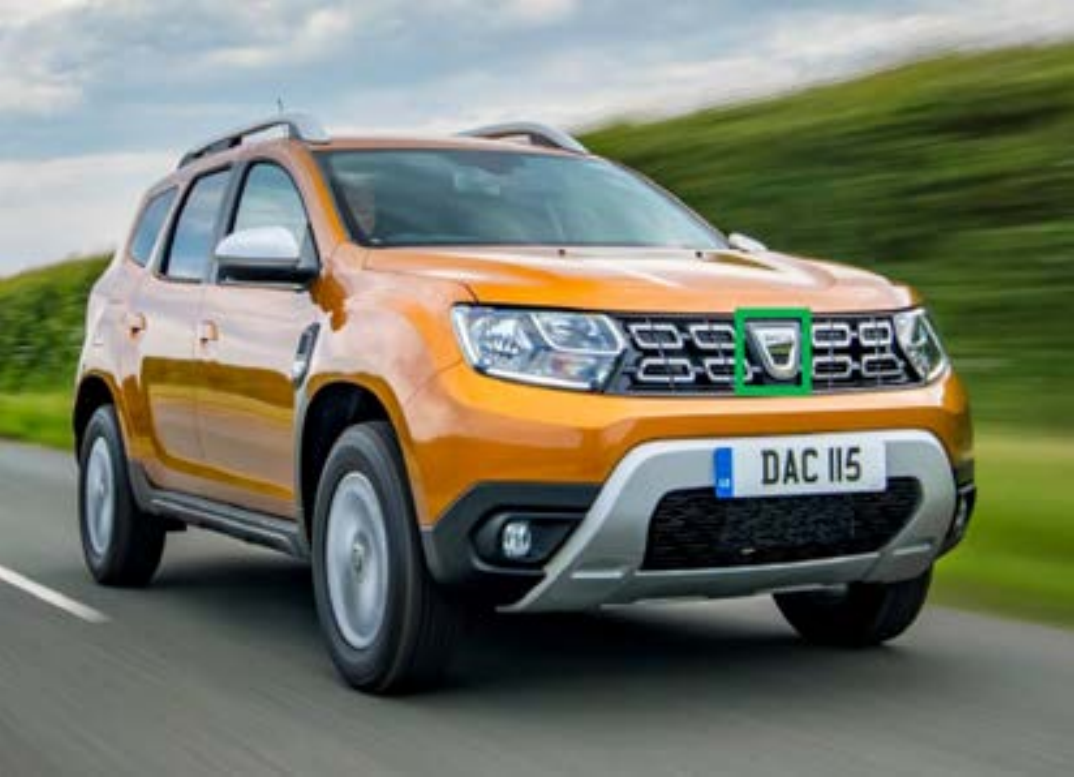}
		%\caption{fig1}
	}
	\hspace{-3mm}
    \subfigure[multi-scale logos]{
    
        \includegraphics[width=2.8cm]{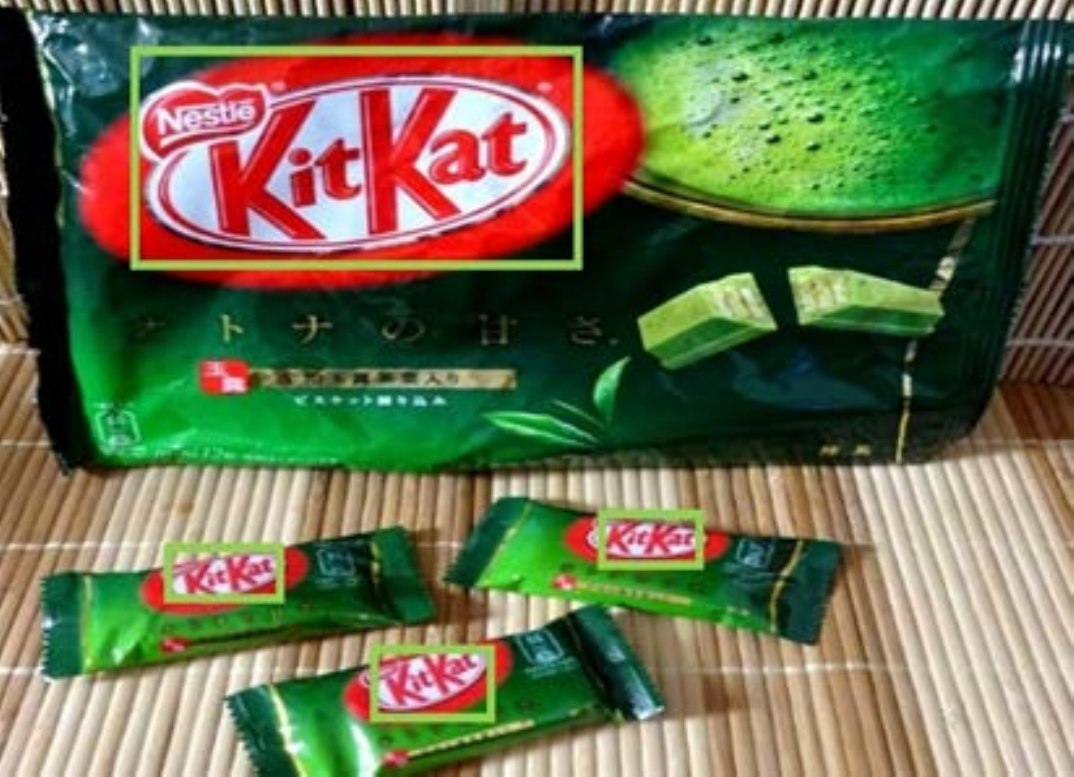}
        %\caption{fig1}
	}
	\hspace{-4mm}
	\subfigure[non-rigid deformation]{
	
        \includegraphics[width=2.8cm]{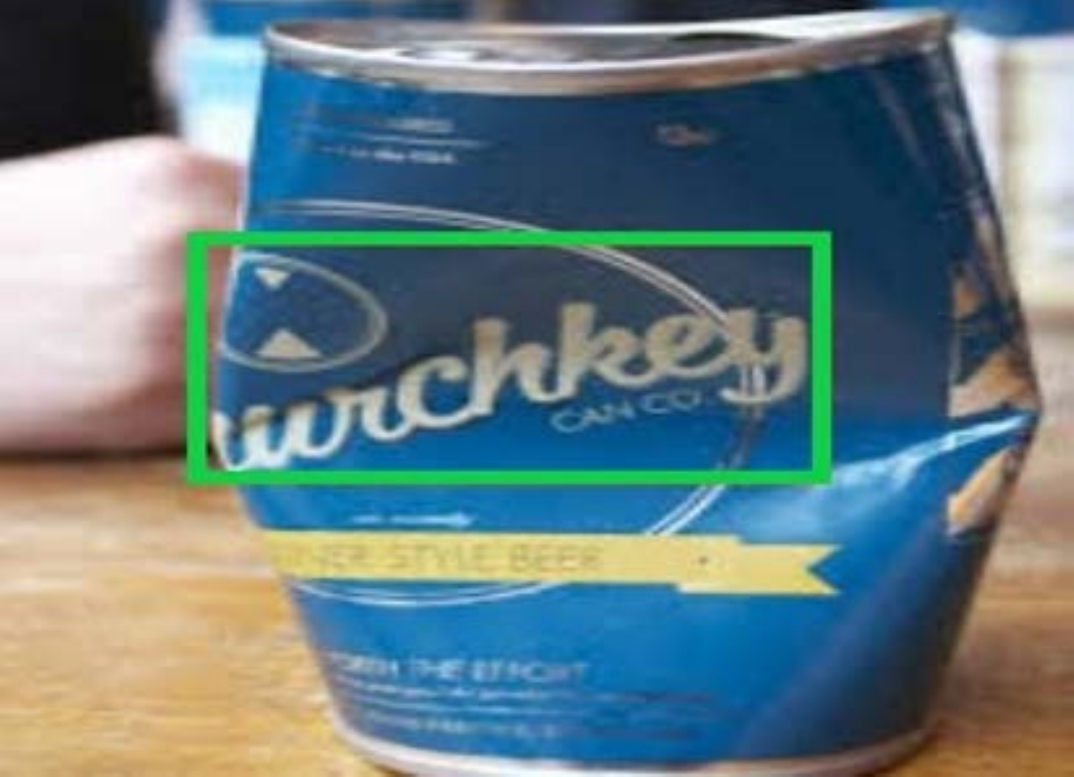}
        %\caption{fig1}
	}
	\hspace{-4mm}
	\subfigure[glare reflection]{
	
        \includegraphics[width=2.8cm]{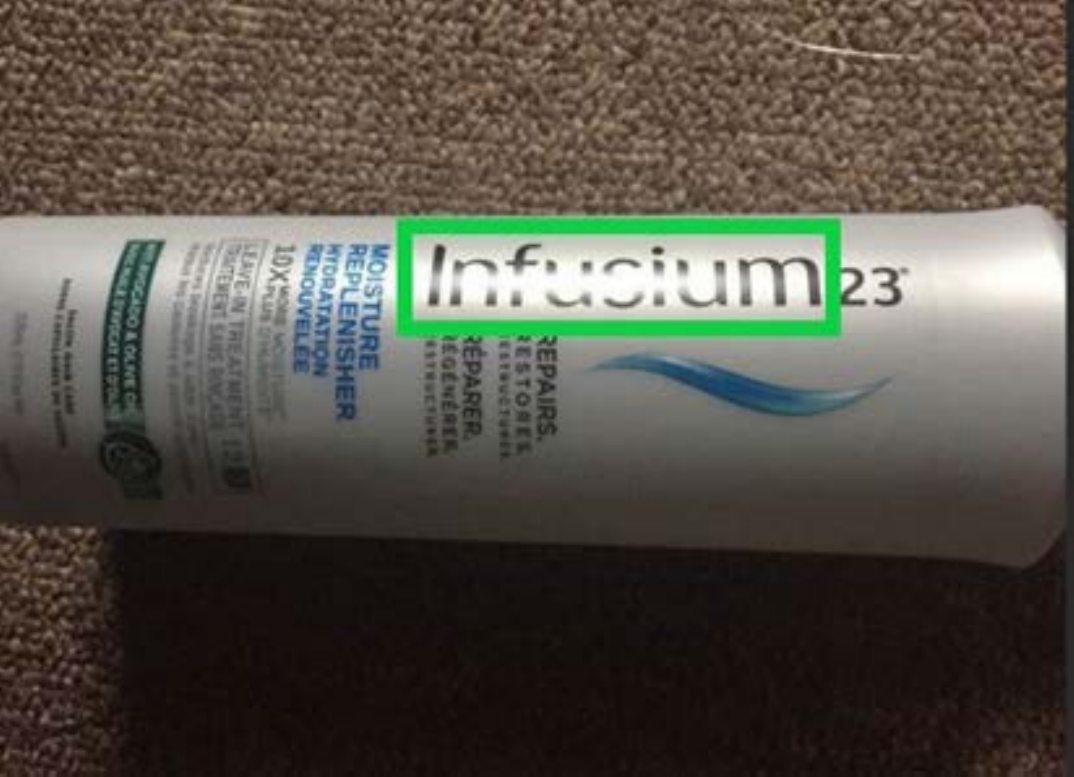}
        %\caption{fig1}
	}
	\caption{Examples of images in adverse conditions.}
	%\label{Fig: O2O_RF_MCC5}	
 \end{figure}
 
Many previous works on logo detection employ hand-crafted features (like SIFT~\cite{sahbi2012context}) to represent logos and use statistical classifiers for classification. Such methods suffer from complex image preprocessing pipelines and poor robustness when dealing with a much larger number of logos. Recent years have witnessed the rousing success of deep learning since ImageNet Large Scale Visual Recognition Challenge (ILSVRC)~\cite{Krizhevsky2012ImageNetCW}. Deep learning-based solutions with expressive feature representation capability offer better robustness, accuracy, and speed and thus attract increasing attention. There are more than 100 papers about logo detection from 1993 to 2022, and a concise milestone of logo detection is shown in Fig. 2. We can see that many deep learning-based logo detection strategies have been proposed since 2015. This survey mainly concentrates on deep learning-based solutions specially developed for logo detection. 
\begin{figure*}[htbp]
		\centerline {\includegraphics[width=18cm]{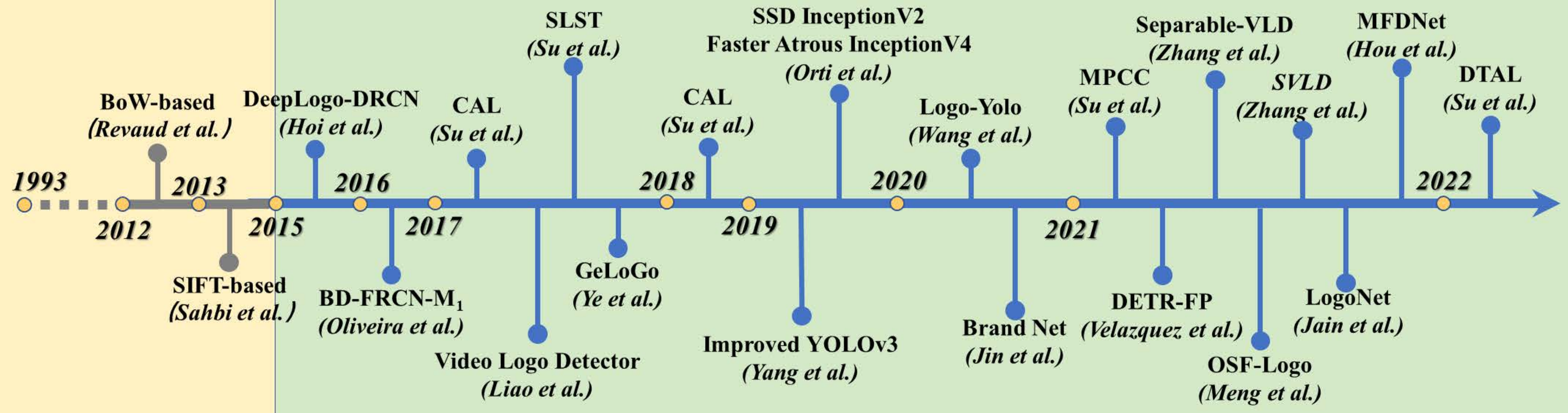}}
		\caption{ A concise milestone for logo detection.}	
		\label{figLabel}
\end{figure*}

Even though deep learning has dominated the logo research community, a comprehensive and in-depth survey on deep learning-based solutions is lacking. In this survey, we mainly focus on the advances in recent deep learning for logo detection. We provide in-depth analysis and discussion on existing studies in various aspects, covering datasets, pipelined used, task types, detection strategies, loss functions, their contributions and limitations. We also try to analyze potential research challenges and future research directions for logo detection. We hope our work could provide a novel perspective to promote the understanding of deep learning-based logo detection, foster research on open challenges, and speed up the development of the logo research field. 

The rest of this paper is organized as follows. In Section II, we investigate the public logo detection datasets. In Section III, we review and organize the currently available work on logo detection. In Section IV, we introduce the applications of logo detection in real-world scenarios. In Section V, we discuss its challenges and prominent future research directions. Finally, we summarize the whole text in Section VI.

\begin{figure*}[htbp!]  
	\centering
	\subfigure[BelgaLogos]{
		\includegraphics[width=4.2cm]{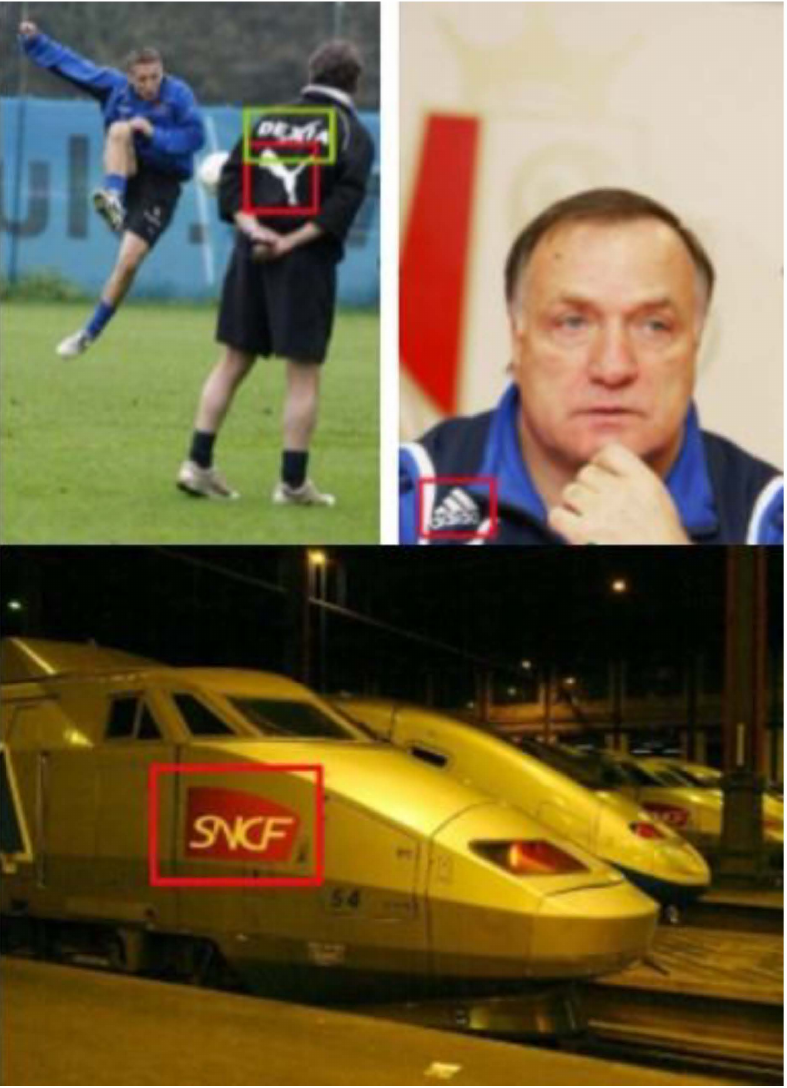}
		%\caption{fig1}
	}	
	%	\quad
	\subfigure[FoodLogoDet-1500]{
		\includegraphics[width=4.2cm]{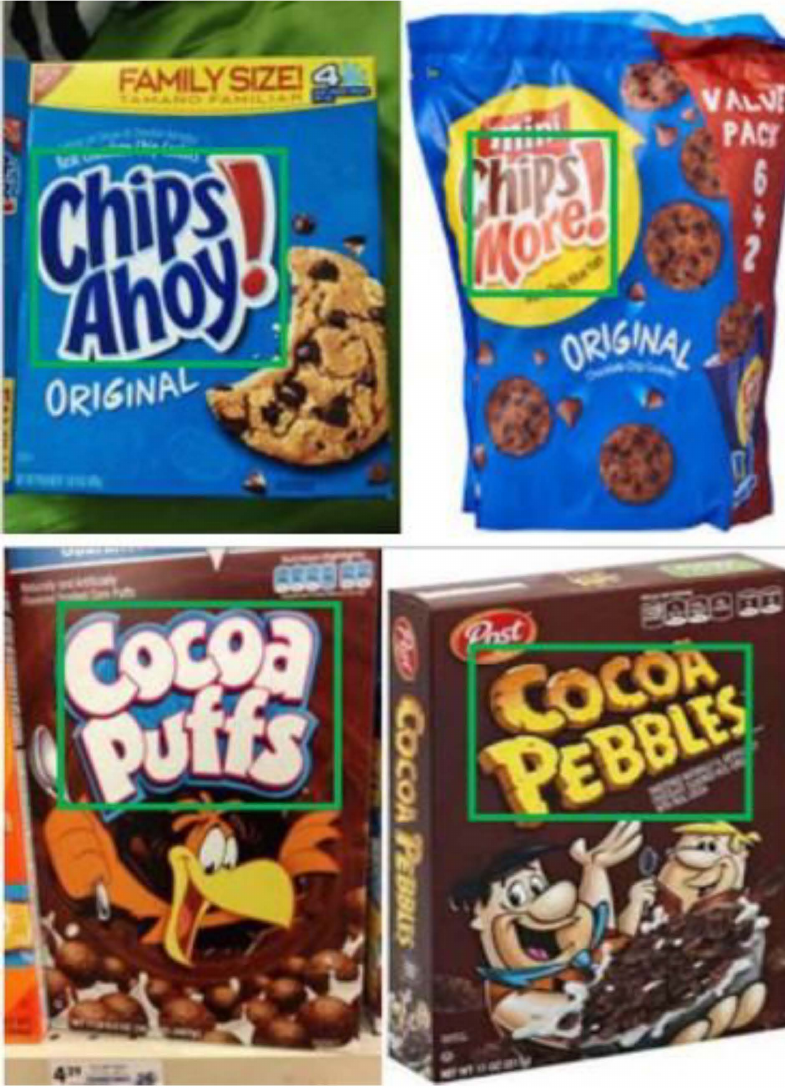}
	}	
	%	\quad
	\subfigure[QMUL-OpenLogo]{
		\includegraphics[width=4.2cm]{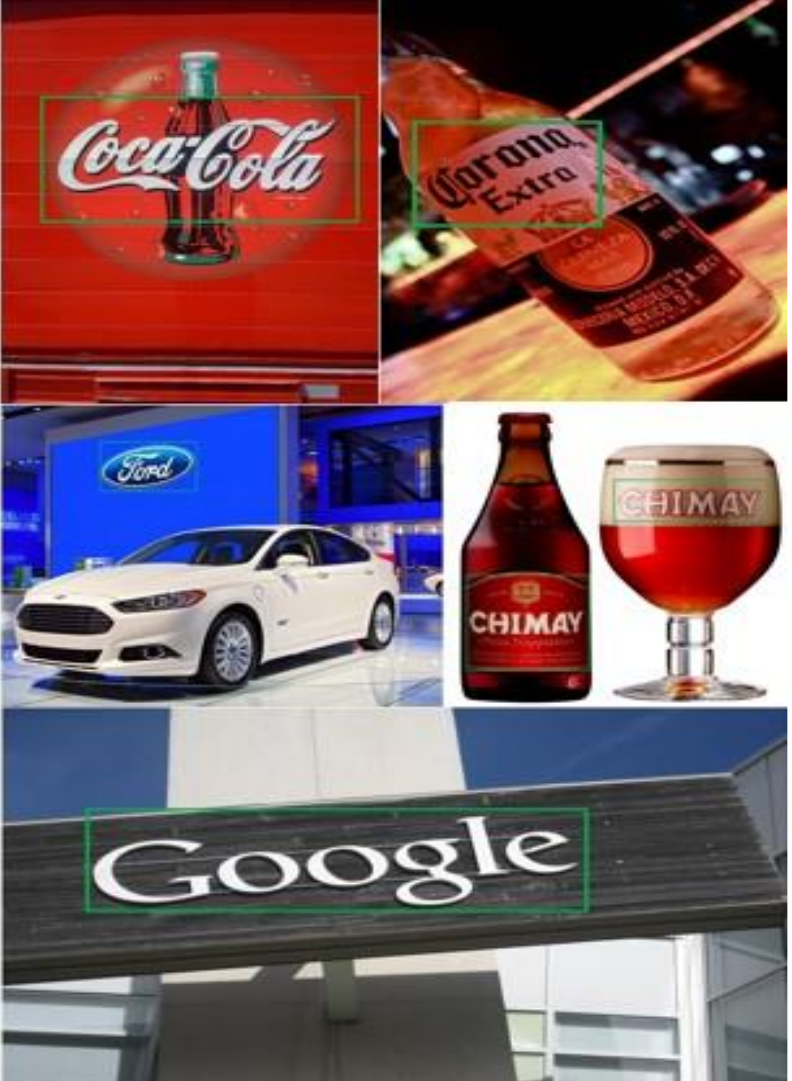}
	}	
	%	\quad
	\subfigure[LogoDet-3K]{
		\includegraphics[width=4.2cm]{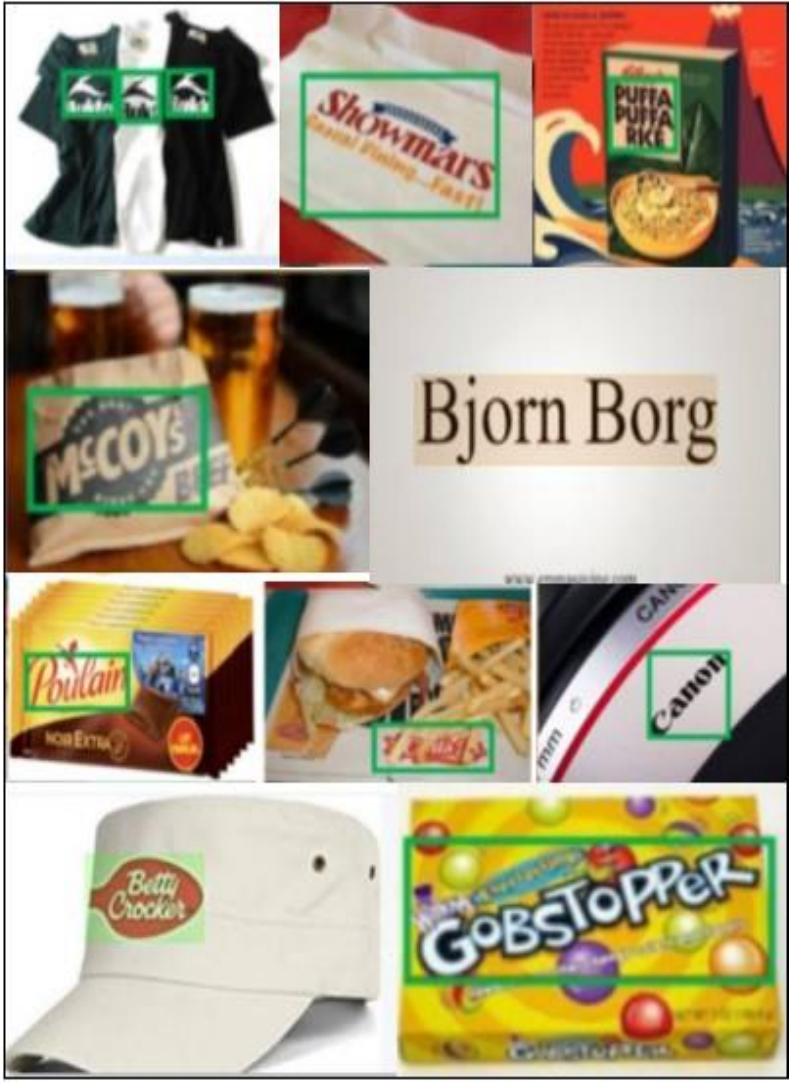}
	}
	\caption{Logo images sampled from different datasets.}
	\label{Fig: O2O_RF_MCC5}	
\end{figure*}

\section{Logo Datasets}

Deep learning has brought great success to object detection in recent years, where datasets play a crucial role. Datasets are not only a common basis for comparing and measuring the performance of algorithms but also an essential factor in supporting advanced object detection algorithms. This section provides an overview of the logo datasets for detection. In addition, we also give a brief summary of existing logo classification datasets.

In recent years, many logo datasets intended for detection have been created to solve the problem of large realistic datasets with accurate ground truth. The use of these datasets enables qualitative as well as quantitative comparisons and allows benchmarking of different algorithms. We conducted statistics on existing datasets commonly used for logo detection and classified them into three types based on their scale: small-scale, medium-scale, and large-scale. Table I provides statistics on the available logo datasets, and Fig. 3 gives some illustrative examples from these datasets.

The small-scale datasets include BelgaLogos~\cite{17joly2009}, FlickrLogos-32~\cite{romberg2011scalable}, etc.
As the first benchmark dataset proposed for logo detection, BelgaLogos~\cite{17joly2009} consists of 10,000 images of natural scenes, with 37 different logos and 2,695 instances of logos labeled with bounding boxes.
FlickrLogos-32~\cite{romberg2011scalable}, one of the most popular small-scale datasets for logo detection, comprises 32 different classes with 70 images in each class. The images in this dataset are mainly captured from the real world, and many contain occlusions, appearance changes, and lighting changes, making detecting this dataset very challenging.
% VLD-30~\cite{yang2018new} is another small-scale dataset of vehicle logos that consists of 30 car categories, with 67 images per category via web crawler. Since this dataset has a biggish dimensional span, it can be used to improve the generalization ability of the training model very well.

Datasets with medium-scale include Logo-Net~\cite{5hoi2015}, QMUL-OpenLogo~\cite{8su2018open}, FoodLogoDet-1500~\cite{Hou2021FoodLogoDet1500AD}, etc.
Logo-Net~\cite{5hoi2015} is built for detecting logos and identifying brands from real-world product images. It consists of two sub-datasets, namely Logo-18 and Logo-160.
QMUL-OpenLogo~\cite{8su2018open} is an open benchmark for logo detection, constructed by aggregating seven existing datasets and building an open protocol for logo detection evaluation. The QMUL-OpenLogo dataset has a highly imbalanced distribution and significant variation in scale, which is critical to verify the performance of the detection algorithm.
FoodLogoDet-1500~\cite{Hou2021FoodLogoDet1500AD} is the first high-quality public dataset of food logos with uneven distribution among different food logo classes, which poses a challenge to the small sample food logo detection algorithms. The dataset is composed of 1,500 food logo classes with 99,768 images.
% Weblogo-2M\cite{su2017weblogo} is obtained by automatic web data acquisition and processing. The dataset removes images with small widths and/or heights, as well as duplicate images. Compared to other databases, the Weblogo-2m dataset presents three unique properties inhernet to large scale data exploration for learning scalable logo models: (1) Weak Annotation. (2) Noisy (False Positives). (3) Class Imbalance.

There are also some large-scale datasets, such as LogoDet-3K~\cite{4wang2020} and  PL8K~\cite{Li2022SeeTekVL}.
LogoDet-3K~\cite{4wang2020} divides all logos into nine super-classes based on the needs of daily life and the main positioning of common enterprises, namely Clothing, Food, Transportation, Electronics, Necessities, Leisure, Medicine, Sports, and Others. The dataset consists of 3,000 logo classes, 158,652 images, and 194,261 logo objects. The number of logo classes contained in different super-classes varies greatly. For example, the “Food”, “Clothes”, and “Necessities” contain more images and objects than other super-classes. The imbalanced distribution across different logo classes of LogoDet-3K poses a challenge to effectively detecting logos with few samples.
PL8K~\cite{Li2022SeeTekVL} is a large logo detection dataset constructed semi-automatically. The dataset consists of 7,888 logo brands and 3,017,146 images, and at least 20 images per class.

\begin{table*}[!t]
\caption{Statistics of existing logo detection datasets.}
\label{Logo_dataset_compare}
\centering
\setlength{\tabcolsep}{5mm}{
\footnotesize
\begin{tabular}{cccccccc}
\hline
\#Scale&\#Datasets&\#Logos&\#Brands & \#Images& \#Objects & \#Public&\#Year \\ \hline
\multirow{11}{*}{Small}&BelgaLogos~\cite{17joly2009}&37&37& 10,000&2,695  & Yes& 2009\\

\multirow{11}{*}{}&FlickrLogos-27~\cite{kalantidis2011scalable}&27&27&1,080&4,671&Yes&2011\\

\multirow{11}{*}{}&FlickrLogos-32~\cite{romberg2011scalable}&32&32&2,240& 5,644 & Yes&2011\\

\multirow{11}{*}{}&MICC-Logos~\cite{sahbi2012context} &13&&720&- &No&2013 \\

\multirow{11}{*}{}&Logo-18~\cite{5hoi2015}&18&10&8,460& 16,043& No&2015\\

\multirow{11}{*}{}&Logos-32plus~\cite{bianco2017deep}&32&32&7,830&12,302 & No& 2017 \\

\multirow{11}{*}{}&Top-Logo-10~\cite{su2017deep} & 10   & 10  & 700& -   & No  & 2017 \\

\multirow{11}{*}{}&Video SportsLogo~\cite{liao2017mutual}&20&20& 2,000&  -    & No &2017 \\

\multirow{11}{*}{}&VLD 1.0~\cite{19liu2019large}&66&66&25,189&-&No&2019\\

\multirow{11}{*}{}&SportLogo~\cite{kuznetsov2020new}&31&31&2,836&-&Yes&2020\\

\multirow{11}{*}{}&VLD-45~\cite{3zhang2021}&45&45&45,000&-&No&2020\\\hline

\multirow{5}{*}{Medium}&Logo-160~\cite{5hoi2015}&160&100&73,414& 130,608   & No &2015   \\

\multirow{5}{*}{}&Logos-in-the-Wild~\cite{20andras2017}&871&871&11,054&32,850& Yes&2017\\

\multirow{5}{*}{}&QMUL-OpenLogo~\cite{8su2018open}&352&352&27,083&-  & Yes&2018 \\

\multirow{5}{*}{}&PL2K~\cite{fehervari2019scalable}&2,000&2,000& 295,814&-& No&2019 \\

\multirow{5}{*}{}&FoodLogoDet-1500~\cite{Hou2021FoodLogoDet1500AD}&1,500&-&99,768&145,400 &Yes&2021\\\hline

\multirow{3}{*}{Large}&Open Brands~\cite{7jin2020open}&1,216&559&1,437,812&3,113,828 &No&2020 \\

\multirow{3}{*}{}&LogoDet-3K~\cite{4wang2020}&3,000&2,864&158,652&194,261&Yes&2020  \\

\multirow{3}{*}{}&PL8K~\cite{Li2022SeeTekVL}&7,888&7,888&3,017,146&-&No&2022\\
 \hline
\end{tabular}}
\end{table*}

In addition, there are also some datasets built for logo classification, e.g. WebLogo-2M~\cite{su2017weblogo} and Logo 2K+~\cite{1wang2020logo}. Weblogo-2M~\cite{su2017weblogo} is obtained by automatic web data acquisition and processing. The dataset excludes images with small widths and/or heights and duplicate images. Compared with other datasets, the Weblogo-2M presents three unique properties inherent to large-scale data exploration for learning scalable logo models: (1) Weak Annotation. (2) Noisy (False Positives). (3) Class Imbalance. Logo-2K+~\cite{1wang2020logo} is a large-scale high-quality logo dataset. It covers a variety of logo classes from the real world, and different types of logo images have various logo appearances, scales, and backgrounds since they are collected from different websites. The dataset has high spatial coverage of categories including Food, Clothes, Institutions, Accessories, Transportation, Electronics, Necessities, Cosmetics, Leisure, and Medical. The number of images is imbalanced among different categories. For instance, “Food” has 769 logo classes, while “Medical” has only 48.

As with common object detection, mAP~\cite{mapiou} is the most commonly used evaluation metric to measure logo detectors.

\begin{figure*}[htbp]
		\centerline {\includegraphics[width=18cm]{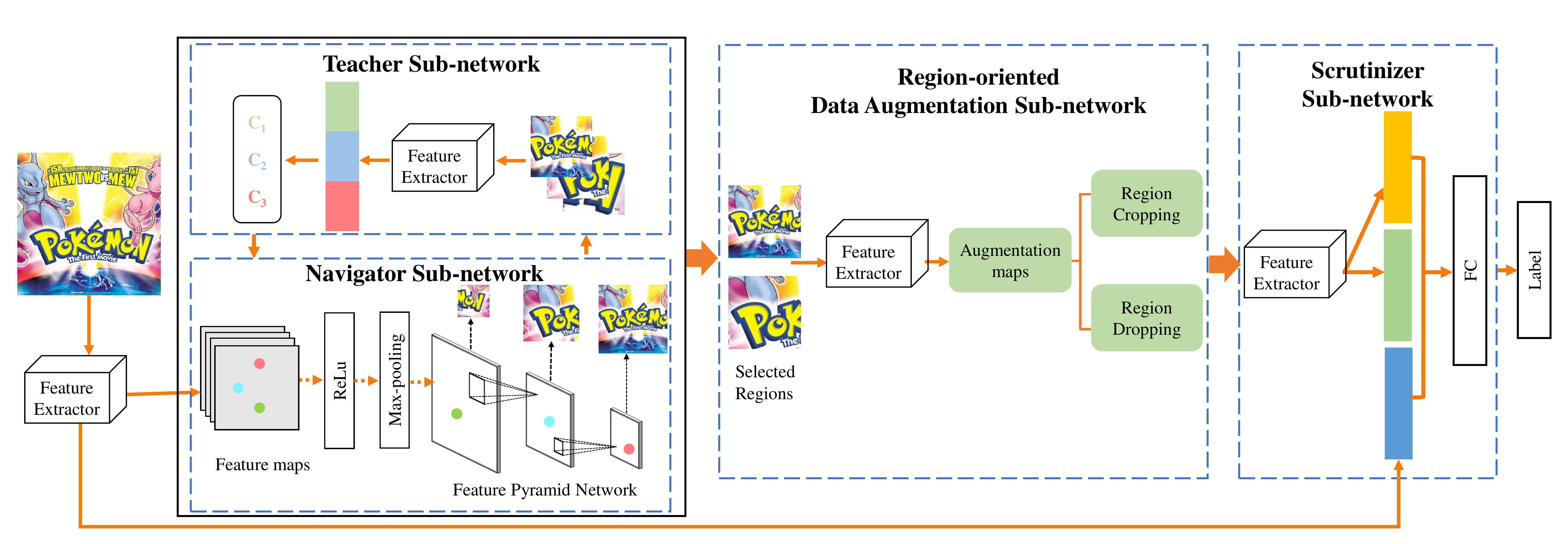}}
		\caption{The architecture of DRNA-Net~\cite{1wang2020logo}.}	
		\label{figLabel}
\end{figure*}

\section{Logo Detection}
In logo detection, logo classification is also an essential part. Therefore, we briefly summarize logo classification before introducing logo detection.
\subsection{Logo Classification}
As one of the most critical tasks in computer vision, logo classification aims to recognize the logo name corresponding to the input image. According to different feature extraction strategies, existing classification methods are divided into two categories: \emph{traditional machine learning-based methods} and \emph{deep learning-based methods}. Some representative methods will be described briefly in the following. 

\subsubsection{Traditional Machine Learning-based Methods}
The traditional logo classification methods extract features through manual features, such as SIFT and HOG, and then classify them by a classifier. 
Support Vector Machine (SVM)~\cite{SVM} is a classifier that performs binary data classification in a supervised learning manner. In traditional logo classification, SVM also shows excellent performance~\cite{7Lei2012,17Zhang2013,carvalho2021automatic,sathiaseelan2021logo}.
Carvalho \emph{et al}.~\cite{carvalho2021automatic} proposed a self-learning and automatic detection method that performs detection without any prior data. The scheme automatically identifies the candidate regions of the localization logo, uses the HOG features extracted from the localization to train the object detector and some sub-detectors and uses the SVM to recognize the logo image.

K-Nearest Neighbor (KNN) is a supervised learning algorithm commonly used in classification~\cite{23xiang2016,15gopinathan2018}.
Gopinathan \emph{et al}.~\cite{15gopinathan2018} proposed a vehicle logo recognition system in 2018. They used Euclidean distance on the initial training dataset to summarize the pixel intensity distribution by applying each channel's mean and standard deviation and used the K-means algorithm to cluster the color histogram features to group different logos. Then HOG and KNN algorithms extract logo features and classify logos.

\subsubsection{Deep Learning-based Methods}

In recent years, with the continuous development of deep learning, deep learning based solutions have also been successfully applied to logo classification~\cite{22pan2013,21zhang2017tv,Hou2017DeepHR,2Gallego2019,20karimi2019,Bernabeu2022MultiLabelLR,13wang2018,A2015DeepLogoHL}.
Karimi \emph{et al}.~\cite{20karimi2019} used the DCNN logo recognition algorithm, which conducted a pre-trained model for feature extraction and then used SVM for logo classification. They also used transfer learning to improve existing pre-trained models for logo recognition. Finally, the fine-tuned deep model is applied to the parallel structure to obtain a more efficient deep model for logo recognition.

The latest trend in logo classification is to design efficient networks with limited resources~\cite{1wang2020logo,2021Patch,lu2021category,Li2022SeeTekVL}.
Wang \emph{et al}.~\cite{1wang2020logo} proposed a Discriminative Region Navigation and Augmentation Network (DRNA-Net), which is capable of discovering more informative regions and expanding these regions for logo classification. DRNA-Net is mainly divided into four parts: navigator sub-network, teacher sub-network, region-oriented data augmentation sub-network, and scrutinizer sub-network. Specifically, the navigation sub-network first computes the amount of information in all regions generated by pre-defined anchors in the image. In order to make the navigation sub-network selects the most informative logo-relevant regions, the teacher sub-network then evaluates each region’s confidence that belongs to the ground-truth class. The region-oriented data augmentation sub-network can augment the selected regions to localize more relevant logo regions. Finally, the features of augment regions and the whole image are fused by the scrutinizer sub-network to obtain a unified feature representation for logo prediction. The architecture of DRNA-Net is shown in Fig. 4. 
In order to distinguish visually similar logos, Li \emph{et al}.~\cite{Li2022SeeTekVL} proposed a multi-task learning architecture named SeeTek based on deep learning and scene text recognition. This model combines deep metric learning and text recognition into a single model leading to considerable performance improvements. The architecture of SeeTek is shown in Fig. 5.

\subsection{Logo Detection}
Logo detection can be seen as a particular case of general object detection. The purpose of logo detection is to detect logo instances of predefined logo classes in images or videos~\cite{5hoi2015}. In this section, we will present the current state-of-the-art work on logo detection in detail. 

Before the development of deep learning, early approaches to logo detection were implemented based on hand-crafted visual features (e.g., SIFT and HOG) and traditional classification models (e.g., SVM)~\cite{t1,t2,t3,t5}. For example, Sahbi \emph{et al}.~\cite{sahbi2012context} proposed a logo matching system in 2013, where they used SIFT to conduct logo detection and recognition. The designed system can not only recognize and match multiple reference logos in an image but also is suitable for detecting similar logos. However, such traditional logo detection methods have certain limitations: (1) The region-selective search algorithm based on sliding windows lacks pertinence, especially with high time complexity. (2) The hand-designed features lack robustness to the variation of logo diversity.

In recent research, deep learning has emerged as the mainstream for logo detection. According to different learning strategies, we categorize these models into region-based Convolutional Neural Network models, YOLO-based models, Single Shot Detector-based models, Feature Pyramid Network-based models and other models.

\begin{figure*}[htbp]
		\centerline {\includegraphics[width=16.5cm]{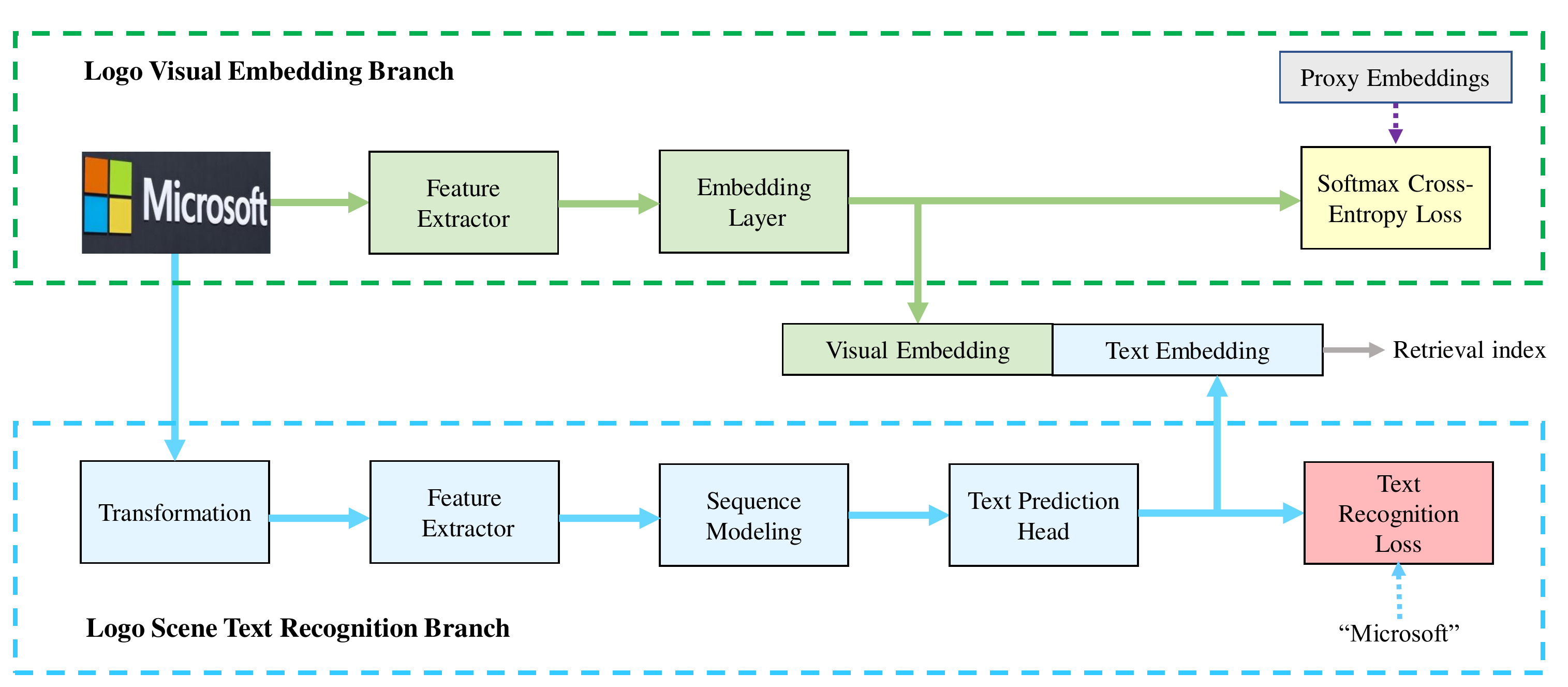}}
		\caption{The architecture of SeeTek~\cite{Li2022SeeTekVL}.}	
		\label{figLabel}
\end{figure*}

\subsubsection{Region-based Convolutional Neural Network models (RCNNs)}

Region-based Convolutional Neural Network (\textbf{R-CNN}~\cite{RCNN}) is a typical region proposals-based approach. Compared with traditional detection algorithms, R-CNN has achieved considerable performance improvement. However, R-CNN still has several shortcomings. For example, R-CNN uses the Selective Search (SS)~\cite{ss} algorithm to generate many overlapping boxes for redundant computation, leading to slow detection speed and taking up a large amount of storage space. The later proposed \textbf{Fast R-CNN}~\cite{fastrcnn} solves this issue by creating an end-to-end trainable system. However, Fast R-CNN still uses the SS algorithm to generate region proposals, limiting its speed. To solve this problem, Ren \emph{et al}. proposed \textbf{Faster R-CNN}~\cite{Ren2015FasterRT} in 2015, which adopts region proposal network (RPN) to generate region proposals. Although Faster R-CNN has advantages in speed and efficiency, it still suffers from computational redundancy and low efficiency of small object detection.

Recently, RCNN-based logo detection has also achieved considerable success~\cite{20andras2017,new1,Orti2019RealTimeLD,sahel2021logo}.
Since the same brand may contain multiple products, different product images of the same brand may present different visual content, and traditional image recognition methods cannot directly solve the problem of brand recognition. Hoi \emph{et al}.~\cite{5hoi2015} proposed a DeepLogo-DRCN scheme for logo detection and brand recognition by exploring several deep region-based convolutional network (DRCN) techniques for object detection. They adopt Selective Search~\cite{ss} to generate regions of interests (RoIs) and input the RoIs into a fully convolutional network (FCN). Feature vectors are then obtained through fully connected layers (FCs) to train the final classifiers and bounding box regressors. The overall architecture is trained in an end-to-end way. This is the first work on deep learning-based logo detection, which provides a new perspective on logo detection research.

We know that training a large-scale logo detection model requires a large amount of data, and the logo datasets usually suffer from data scarcity. Data augmentation and transfer learning are common solutions to release the problem of data scarcity. 
Oliveira \emph{et al}.~\cite{oliveira2016automatic} proposed an automatic graphic logo detection system based on Fast R-CNN~\cite{fastrcnn}, which is robust to unconstrained imaging conditions. They used transfer learning and data augmentation to train a convolutional neural network model and allowed multiple detections of potential regions containing objects. Experimental results demonstrate that this strategy is superior to the traditional methods~\cite{romberg2011scalable}. However, the adopted SS algorithm still generates redundant calculations, reducing detection speed.
In addition, Li \emph{et al}.~\cite{li2017graphic} built Faster R-CNN for logo detection by using transfer learning, data augmentation, and clustering to guarantee suitable hyperparameters and more precise anchors for RPN. The experimental results show that this improvement could significantly improve the detection accuracy.

\subsubsection{YOLO-based models (YOLOs)}
YOLOs are single-stage detectors that are commonly used for logo detection. YOLOs directly detect images and outputs the category and location information of the detected objects. Early YOLO~\cite{yolo1} has fast detection speed, but at the expense of accuracy, especially for small objects. The later improvements, like YOLOv2~\cite{yolo2},  YOLOv3~\cite{yolo3}, YOLOv4~\cite{yolo4}, YOLOF~\cite{yolof}, YOLOR~\cite{yolor}, YOLOX~\cite{YOLOX}, YOLOv6~\cite{YOLO6}, YOLOv7\cite{yolo7} improve the overall performance of YOLO, balancing the accuracy and speed.
% \textbf{YOLO}~\cite{yolo1}, proposed by Joseph \emph{et al}. in 2015, is the first single-stage detector. The input image is divided into a grid of \textit{s × s}. It will be responsible for predicting the object if the center of an object falls in this grid. In addition to the location of the bounding box, each set of predictions also needs to predict a confidence score to represent the IoU between the bounding box and the ground truth box. YOLO has a fast detection speed, but there are some shortcomings in accuracy, especially for small objects. The later proposed YOLOv2~\cite{yolo2} improves the overall performance of YOLO, balancing the accuracy and speed. In recent years, more YOLO-based improvements~\cite{yolo3,yolo4,yolof,yolor,YOLOX} are proposed.

YOLOs are widely used in vehicle logo detection. 
Given that real-time and high efficiency is essential issues in vehicle logo detection. Inspired by the superiority of deep learning in feature extraction, Yin \emph{et al}.~\cite{yin2020real} proposed a high-efficiency vehicle logo detection system based on YOLOv2~\cite{yolo2}. Compared with the traditional methods based on manual extraction features, the system has the advantages of self-learning features and direct image input and can realize the dual functions of positioning and recognition of the vehicle logo. 
In the case of complex backgrounds, a vehicle logo usually occupies only a small part of the image. Therefore, it is challenging to identify vehicle logos in real-world scenarios accurately.
%The detection of small objects by single-stage detectors is often challenging, since sliding windows will miss some small objects. The main reason for the difficulty of small object detection is that the features extracted from shallow layers are not obvious, and the features extracted from deep layers are easy to cause object loss. Therefore, a suitable feature extraction network is necessary. 
To solve this issue, Yang \emph{et al}.~\cite{yang2019fast} proposed a data-driven enhanced training method based on YOLOv3~\cite{yolo3}. They combined a feature extraction network with a multi-scale decision scheme to improve the detection accuracy of vehicle logos. However, the detection results of vehicle logos in complex scenes are unsatisfactory.
Based on \cite{yang2019fast}, Zhang \emph{et al}.~\cite{zhang2021vehicle} analyzed the characteristics of vehicle identification in static images and found that both the network for feature extraction and the training strategy for detection has a significant impact on the accuracy of vehicle identification. They proposed a lightweight network structure with separable convolutions to replace traditional methods to improve detection accuracy and speed under the YOLO-based framework. Experiments show that this method improves the real-time performance of vehicle logo detection and also improves the detection accuracy of small-scale objects while maintaining the speed to a certain extent.

In various fields, the frequency of different logos usually varies widely. Therefore, the number of different logo classes is often unbalanced, making it difficult for models to detect logo classes with small samples correctly. Wang \emph{et al}.~\cite{4wang2020} proposed a robust baseline method Logo-Yolo based on YOLOv3~\cite{yolo3}. 
They first adopt the K-means clustering algorithm to select the number of candidate anchor boxes and aspect ratio dimensions. They then used Focal loss to address the imbalance of samples in the dataset. Lastly, they introduced CIoU loss to improve bounding box regression results. Compared with YOLOv3~\cite{yolo3}, the performance of this method is significantly improved in predicting small objects and objects in complex backgrounds. However, in some specific cases, the method also shows poor performance, such as detecting similar or occluded logos.

\begin{figure*}[htbp]
		\centerline {\includegraphics[width=18cm]{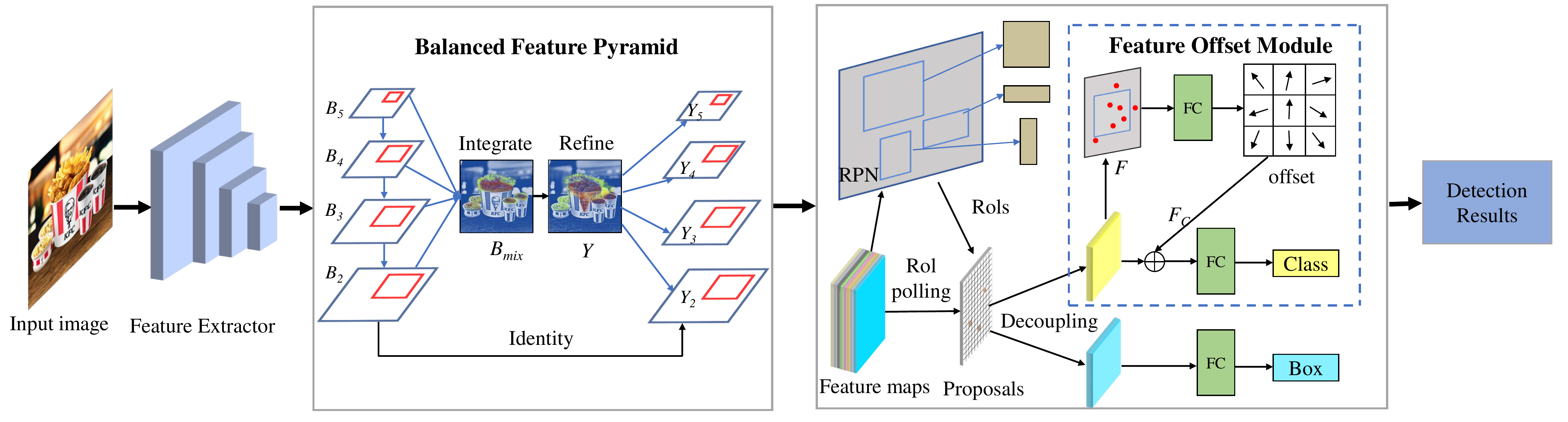}}
		\caption{ The architecture of MFDNet~\cite{Hou2021FoodLogoDet1500AD}.}	
		\label{figLabel}
\end{figure*}

\subsubsection{Single Shot Detector-based models (SSDs)}
Single Shot MultiBox Detector (SSD)~\cite{ssd} is another single-stage detector using a single deep neural network. SSD uses multi-scale feature maps to detect objects at different scales, with the bottom layer predicting small objects and the top layer predicting large objects. While ensuring the detection speed, the detection accuracy of SSD outperforms a comparable state-of-the-art Faster R-CNN model. 
% Single Shot MultiBox Detector (SSD)~\cite{ssd} proposed by Liu \emph{et al}. in 2015 is the second single-stage detector. While ensuring the detection speed, the detection accuracy of SSD is comparable to Faster R-CNN. SSD uses multi-scale feature maps to detect objects at different scales, with the bottom layer predicting small objects and the top layer predicting large objects. Due to the various aspect ratios of objects in real life, multiple anchors with aspect ratios are set in SSD to detect objects with different aspect ratios. Although SSD has improved the detection accuracy, it is still insufficient in the detection of small objects.
Therefore, SSDs are also widely used in vehicle logo detection~\cite{Orti2019RealTimeLD}.
% SSD model is also widely used in logo detection. A number of works~\cite{Orti2019RealTimeLD,patalappa2021robust} are based on SSD for improvement. 
Previous works usually detect vehicle logos at the same scale without considering the multi-scale variation of vehicle logos. However, the vehicle logos captured by the camera are of various sizes. It is essential to design an algorithm with multi-scale vehicle logo detection ability. To this end, Zhang \emph{et al}.~\cite{3zhang2021} proposed a multi-scale vehicle logo detection (SVLD) based on SSD~\cite{ssd} to assist in the recognition of vehicles by modifying the pretraining strategy, adding a feature extraction layer, and controlling the ratio of positive and negative samples. According to the characteristics of the vehicle, a preset frame and parameter settings are designed to effectively reduce redundant calculations and accelerate the convergence of the model quickly. Compared with SSD, the detection accuracy of SVLD is improved, but the time complexity is slightly increased due to the addition of the feature extraction layer.
In addition, although SSDs have competitive accuracy compared to other single-stage methods, they are still insufficient in detecting small objects.

%There are also some works that use SSD for incremental logo detection.  The logo detector needs to be retrained when learning new classes, which consumes a lot of resources. Therefore, it is necessary to train a learning system that can learn new classes based on the original classes. RFBNet~\cite{RFBNet} is built on SSD with the backbone network VGG~\cite{VGG}, which combines the ideas of inception and atrous convolution. Bastan \emph{et al}.~\cite{bastan2019large} proposed an open-set logo detection (OSLD) system based on RFBNet. The system can add new logo classes without retraining. Their proposed simple deep metric learning (SDML) framework can effectively improve the performance of OSLD. Experiments on multiple datasets demonstrate that the system has good generalization ability to unseen logo classes.

\subsubsection{Feature Pyramid Network-based models (FPNs)}

Lin \emph{et al}. proposed Feature Pyramid Networks (FPN)~\cite{FPN} in 2017, which mainly solves the multi-scale problem in object detection by changing the connectivity of the network. FPN dramatically improves the performance of small object detection without increasing the amount of calculation. In recent years, many works have adopted FPN to solve problems of logo detection, like multi-scale and small objects~\cite{meng2021adaptive,7jin2020open,velazquez2021logo,Hou2021FoodLogoDet1500AD} and these FPNs have achieved exciting results. 

Aiming at the problems of multi-scale and other geometric variations of logos, Meng \emph{et al}.~\cite{meng2021adaptive} proposed obtaining sufficient features for logo (OSF-Logo) based on FPN. Specifically, they introduced the Regulated Deformable Convolution (RDC) module in a specific layer of FPN, which allows the sampling point positions of the convolution kernel at different positions to adaptively change according to the content, thereby adapting to the geometric changes of the logo. Meanwhile, they adopt an up-sampling operator in FPN to generate an adaptive kernel that fully aggregates rich semantic information. Therefore, OSF-Logo can quickly perceive a wide range of content and effectively detect multi-scale logos.
Similarly, Jin \emph{et al}.~\cite{7jin2020open} designed a network called “Brand Net”. Brand Net uses FPN to extract multi-scale features. On this basis, it introduces an anchor refinement network in FPN to remove negative anchors, thus greatly reducing the number of anchor boxes. Therefore, it can effectively detect logos with different scales and significantly improve the detection performance of small logos.
Detection Transformers (DETR)~\cite{DETR} has problems detecting small objects, so it cannot be directly used for multi-scale logo detection. To solve this issue, Velazquez \emph{et al}.~\cite{velazquez2021logo} incorporated Feature Pyramid (FP) into DETR architecture. They firstly split the penultimate layer of the feature pyramid into four patches of the same size as the smallest feature map in the pyramid. They then fed these patches into the DETR pipeline for prediction. This method effectively detects small objects. However, the backward propagation computation of this method increases memory consumption.
Recently, Hou \emph{et al}.~\cite{Hou2021FoodLogoDet1500AD} proposed a multi-scale feature decoupling network (MFDNet) for logo detection to solve the problem of distinguishing multiple logo categories. MMFDNet comprises two components: Balanced Feature Pyramid (BFP) and Feature Offset Module (FOM). The former is designed to fuse multi-scale features using the same deep feature map to integrate balanced semantic features. At the same time, the latter comprises an automatically learned anchor region proposal network for pixel-level offsets to search for the best features for logos. The architecture of MFDNet is shown in Fig. 6.

\begin{table*}[!t]
\begin{center}
\caption{Summary of representative deep learning-based logo detection methods.}
% \label{Logo_dataset_compare}
% \centering
\setlength{\tabcolsep}{1.7mm}{
\scriptsize
\begin{tabular}{|c|c|c|c|c|c|c|c|}
\hline
\textbf{Method type}&\textbf{Method}& \textbf{Pipeline}& \textbf{Task}& \textbf{Strategy}&\textbf{Loss function}&\textbf{Year}\\
\hline
\multirow{18}{*}{RCNNs}&DeepLogo-DRCN~\cite{5hoi2015}&\begin{tabular}[c]{@{}c@{}}RCNN\\Fast RCNN\\ SPPnet\end{tabular}&Regular logo detection&-&-&2015\\\cline{2-7}
 
\multirow{18}{*}{}&BD-FRCN~\cite{oliveira2016automatic}&Fast RCNN& Regular logo detection&Transfer learning&-& 2016\\\cline{2-7}

\multirow{18}{*}{}&Improved Faster RCNN~\cite{new1}& Faster RCNN& Small logo detection&Preset anchor boxes&-&\multirow{6}{*}{2017}\\\cline{2-6}

\multirow{18}{*}{}&\begin{tabular}[c]{@{}c@{}}Deep Region-based \\Convolutional Networks~\cite{li2017graphic}\end{tabular}&Faster RCNN&Regular logo detection&\begin{tabular}[c]{@{}c@{}}Transfer learning\\K-means clustering\\Data augmentation \end{tabular}&-&\multirow{7}{*}{}\\\cline{2-6}

\multirow{18}{*}{}&Video logo detector~\cite{liao2017mutual}&Fast RCNN&Video logo detection&\begin{tabular}[c]{@{}c@{}}Homogeneity enhancement\\ Mutual enhancement\end{tabular} &-&\multirow{7}{*}{}\\\cline{2-7}

\multirow{18}{*}{}&Faster RCNN+ResNet-50~\cite{chen2021robust}&Faster RCNN &\begin{tabular}[c]{@{}c@{}}Long tail logo detection \\Small logo detection\end{tabular}&\begin{tabular}[c]{@{}c@{}}Data augmentation\\ Category balance\end{tabular}&\begin{tabular}[c]{@{}c@{}}EQLv2\\ Seesaw Loss\\\end{tabular}&\multirow{9}{*}{2021}\\\cline{2-6}

\multirow{18}{*}{}&Cascade detectoRS~\cite{jia2021effective}&Cascade RCNN &\begin{tabular}[c]{@{}c@{}}Long tail logo detection \\Small logo detection\end{tabular}&\begin{tabular}[c]{@{}c@{}}Data augmentation\\Multi-scale training\end{tabular}&EQLv2&\multirow{10}{*}{}\\\cline{2-6}

\multirow{18}{*}{}&\begin{tabular}[c]{@{}c@{}}Cascade RCNN+\\Res2Net101~\cite{leng2021gradient}\end{tabular}&Cascade RCNN &Long tail logo detection &\begin{tabular}[c]{@{}c@{}}Data augmentation\\ Gradient balance\\Data Sampling\end{tabular}&EQLv2&\multirow{10}{*}{}\\\cline{2-6}

\multirow{18}{*}{}&Green Hand~\cite{xu2021simple}&DetectoRS &\begin{tabular}[c]{@{}c@{}}Long tail logo detection\\Small logo detection \end{tabular}&\begin{tabular}[c]{@{}c@{}}Data augmentation\\  Resampling\\Weighted boxes\end{tabular}&EQLv2&\multirow{10}{*}{}\\\hline

\multirow{12}{*}{YOLOs}&Improved YOLOv3~\cite{yang2019fast}&YOLOv3 &Small logo detection &Hard example re-training&-&2019\\\cline{2-7}

\multirow{16}{*}{}&Scene recognition CNN~\cite{kuznetsov2020new}&YOLOv3&Regular logo detection&-&-&\multirow{8}{*}{2020}\\\cline{2-6}

\multirow{16}{*}{}&Improved YOLOv2~\cite{yin2020real}&YOLOv2&Regular logo detection&\begin{tabular}[c]{@{}c@{}}Separable convolution\\Multi-scale features fusion\end{tabular}&\begin{tabular}[c]{@{}c@{}}Confidence error loss\\Coordinate error loss\\Prediction box loss\end{tabular}&\multirow{10}{*}{}\\\cline{2-6}

\multirow{16}{*}{}&Improved YOLOv2~\cite{vehicle}&YOLOv2&Multi-scale logo detection&\begin{tabular}[c]{@{}c@{}}Preset anchor boxes\\K-means clustering\end{tabular}&-&\multirow{10}{*}{}\\\cline{2-6}

\multirow{16}{*}{}&Logo-Yolo~\cite{4wang2020}&YOLOv3 &Regular logo detection&\begin{tabular}[c]{@{}c@{}}Improved losses \\K-means clustering\end{tabular}&\begin{tabular}[c]{@{}c@{}}Focal loss\\CIoU loss\end{tabular}&\multirow{10}{*}{}\\\cline{2-7}

\multirow{16}{*}{}&Scaled YOLOv4~\cite{palevcek2021logo}&YOLOv4 &Regular logo detection &-&-&\multirow{2}{*}{2021}\\\cline{2-6}

\multirow{16}{*}{}&Separable-VLD~\cite{zhang2021vehicle}&YOLO&Small logo detection&Separable convolution&-&\multirow{2}{*}{}\\\hline

\multirow{2}{*}{SSDs}&SVLD~\cite{3zhang2021}&SSD &Multi-scale logo detection&Preset anchor boxes&-&\multirow{2}{*}{2019}\\\cline{2-6}

\multirow{2}{*}{}&SSD InceptionV2~\cite{Orti2019RealTimeLD}&SSD &Regular logo detection&Preset anchor boxes&-&\multirow{2}{*}{}\\\hline

\multirow{8}{*}{FPNs}&Brand Net~\cite{7jin2020open}&-&Small logo detection &\begin{tabular}[c]{@{}c@{}}Soft mask attention\\Weight transfer\\Anchor refinement \end{tabular}&\begin{tabular}[c]{@{}c@{}}Class-aware smooth L1\\Class-agnostic smooth L1\\Cross-entropy loss\end{tabular}&2020\\\cline{2-7}

\multirow{8}{*}{\makecell[c]}&OSF-Logo~\cite{meng2021adaptive}&-&Multi-scale logo detection&\begin{tabular}[c]{@{}c@{}}Deformable convolution\\Up-sampling operator \end{tabular}&-&\multirow{5}{*}{2021}\\\cline{2-6}

\multirow{8}{*}{\makecell[c]}&MFDNet~\cite{Hou2021FoodLogoDet1500AD}&-&Multi-scale logo detection&\begin{tabular}[c]{@{}c@{}}Multi-scale feature fusion\\Deformable learning\end{tabular}&Cross-entropy loss&\multirow{6}{*}{}\\\cline{2-6}

\multirow{8}{*}{\makecell[c]}& DETR-FP~\cite{velazquez2021logo}&DETR&Small logo detection&Model fusion&\begin{tabular}[c]{@{}c@{}}Hungarian loss\\Generalized IoU loss\end{tabular}&\multirow{6}{*}{}\\\hline

\multirow{15}{*}{Others}&SLST~\cite{su2017weblogo}&Faster RCNN& Weakly supervised logo detection&\begin{tabular}[c]{@{}c@{}}Incremental learning\\Self-Training\end{tabular}&-& \multirow{3}{*}{2017} \\ \cline{2-6}

\multirow{18}{*}{}&SCL~\cite{su2017deep}&Faster RCNN&Few shot logo detection &Synthetic context logo&-&\multirow{3}{*}{}\\\cline{2-7}

\multirow{18}{*}{}&CAL~\cite{8su2018open}&\begin{tabular}[c]{@{}c@{}}Faster RCNN\\YOLOv2\end{tabular}&Weakly supervised logo detection &Context adversarial learning&\begin{tabular}[c]{@{}c@{}}Softmax cross-entropy loss\\Conditional adversarial loss\end{tabular}&2018\\\cline{2-7}

\multirow{18}{*}{}&\begin{tabular}[c]{@{}c@{}}Weakly supervised \\learning of CNN~\cite{2020Logo}\end{tabular}&DCNN&Weakly supervised logo detection&\begin{tabular}[c]{@{}c@{}}Generate saliency map\\GrabCut segmentation\end{tabular}&Softmax loss&\multirow{7}{*}{2021}\\\cline{2-6}

\multirow{18}{*}{}&\begin{tabular}[c]{@{}c@{}}Airline logo\\ detection system~\cite{wilms2021airline}\end{tabular}&AttentionMask&Complex scenes logo detection&Data augmentation&Cross-entropy loss&\multirow{10}{*}{}\\\cline{2-6}

\multirow{18}{*}{}&LogoNet~\cite{jain2021logonet}&-&Complex scenes logo detection&Spatial attention&-&\multirow{10}{*}{}\\\cline{2-6}

\multirow{18}{*}{}&Deep attention networks~\cite{yohannes2021domain}&-&Regular logo detection&\begin{tabular}[c]{@{}c@{}}Domain adaptation\\Nonlocal block\end{tabular}&-&\multirow{10}{*}{}\\\cline{2-6}

\multirow{18}{*}{}&MPCC~\cite{su2021multi}&Faster RCNN &Few shot logo detection&\begin{tabular}[c]{@{}c@{}}Domain adaptation\\Data augmentation\end{tabular}&Softmax cross-entropy loss&\multirow{10}{*}{}\\\cline{2-7}

\multirow{18}{*}{}&DTAL~\cite{Su2022FewST}&DCNN&\begin{tabular}[c]{@{}c@{}}Video logo detection\\Few shot logo detection\end{tabular}&\begin{tabular}[c]{@{}c@{}}Active learning\\Transfer learning\end{tabular}&Meta-class loss&2022\\

\hline
\end{tabular}}
\end{center}
\end{table*}

\subsubsection{Other models}
Some other models used in logo detection are introduced in this section.

In order to solve the problem of limited logo data, Su \emph{et al}.~\cite{8su2018open} proposed a data augmentation strategy focusing on logo context optimization: Contextual Adversarial Learning (CAL). CAL takes an image with a logo object as input and generates context-consistent synthetic images that can be used as additional training data. However, the distribution of this synthetic image is different from the actual test image. 
To solve this issue, the same authors proposed a Multi-Perspective Cross-Class (MPCC) domain adaptation method~\cite{su2021multi} based on the baseline method CAL~\cite{8su2018open}. MPCC conducts feature distribution alignment in the data augmentation principle from two perspectives. One is to align the feature distribution between the synthetic logo image of 1-shot icon supervised classes and the authentic logo image of the fully supervised class. The other is to align the feature distribution between the logo and non-logo images. This mitigates the domain shift problem between model training and testing on 1-shot icon supervised logo classes and reduces the model's overfitting to the fully labeled logo classes. The final combination of MPCC with Faster R-CNN achieved good results.

Training a logo detector usually requires a large amount of labeled data, and labeling logo data is time-consuming. Therefore, weakly supervised learning is a vital strategy to solve this issue. Kumar \emph{et al}.~\cite{2020Logo} proposed a method for logo detection by using weakly supervised learning of CNN to generate a deep saliency map, which is generated by a single back-propagation of a trained Deep CNN. The authors also used an interactive Grabcut algorithm to calculate the salient segmented regions of the image. However, this method is not sufficiently modeled to detect multiple brand logos in the same image.

Logo detectors usually perform poorly in classes with a small number of samples. To this end, Yohannes \emph{et al}.~\cite{yohannes2021domain} designed a framework consisting of a domain adaptation that reduces the loss function between source-target datasets and also represents important source features adopted by the target dataset. The authors add nonlocal blocks and attention mechanisms to achieve a decent performance of logo detection.
Similarly, Su \emph{et al}.~\cite{Su2022FewST} developed a deep transfer active learning algorithm named DTAL to select the most valuable samples such that we can label the smallest number of samples to achieve maximum performance improvements in training detection models. 

When the image is affected by unfavorable factors such as illumination, rotation, occlusion, etc., the performance of the detector will be significantly reduced. In recent years, the attention mechanisms have attracted increasing attention to solving this issue in logo detection~\cite{wilms2021airline,jain2021logonet}. For example, Wilms \emph{et al}.~\cite{wilms2021airline} took the airline logos as an example and proposed a logo detection system under adverse conditions. The authors adopt the object proposal generation system AttentionMask~\cite{AttentionMask}. Since the aviation system logo is relatively large, the AttentionMask system removes the scale of small objects to improve execution efficiency and reduce false positives. They also used a data augmentation solution to simulate the effects of severe weather to diversify the training data processing effects. The experimental results show that the proposed strategy can achieve good performance in detecting the logo in complex real-world environments. However, the data obtained under real-world conditions are not as ideal as the data generated by data augmentation, so the performance of the model in detecting images in real-world conditions will be degraded.
Considering that locating logos in complex scenarios is challenging due to the large variety of types and appearance of logos, Jain \emph{et al}.~\cite{jain2021logonet} proposed a logo detection framework called LogoNet, which includes a spatial attention module. LogoNet can capture rich spatial information and focus more precisely on the logo object within an image. LogoNet achieves a significant performance gain within considerable computation time.

In order to analyze current state-of-the-art deep learning-based methods more comprehensively, the representative models are presented in chronological order in Table II, and the experimental results of these models on three popular benchmark datasets, namely Flickrlogos-32, QMUL-OpenLogo, and Open Brands, are listed in Table III, Table IV, and Table V respectively.

\section{Applications}
Logo detection is widely used in various applications, such as intellectual property protection, autonomous driving, and product branding. In this section, we describe the representative applications of logo detection in detail.

\subsection{Brand Monitoring}
The most obvious use case for logo detection is brand monitoring, which includes brand protection, brand recommendation, and brand identification. Logo plays an essential role in business marketing as a unique brand identity. However, there are many types of brands in real life, and there may be subtle logo differences between the two products. Inter-class similarities and intra-class differences will greatly increase the difficulty of detection. In addition, images may have highly diverse context, illumination, projection transformation, and resolution. Therefore, how efficiently detecting brands in images is a challenging task. Currently, there is a lot of work on brand identity~\cite{oliveira2016automatic,Orti2019RealTimeLD,7jin2020open,hu2020multimodal}.
For brand logos, the image usually includes some textual description. Textual information is an important source of information for brand identity. Hu \emph{et al}.~\cite{hu2020multimodal} proposed a multimodal fusion framework for logo detection. The framework combines image-based logo recognition with contextual features for logo detection using a natural language model. Additional contextual information can alleviate the limitations of detection. The performance of this model for logo detection has been improved, but there are still shortcomings in positioning.

\begin{table}[!t]
\begin{center}
\caption{Performance comparison of logo detectors on the FlickrLogos-32 benchmark.}
\begin{math}
\label{Logo_dataset_compare}
% \centering
\setlength{\tabcolsep}{5mm}{
\scriptsize
\begin{tabular}{ccc}
\toprule[1pt]
\textbf{Method}&\textbf{mAP(\%)}&\textbf{Year} \\ 
\toprule[1pt]
FRCN ~\cite{A2015DeepLogoHL}&74.4&2015\\
BD-FRCN-M~\cite{oliveira2016automatic}&73.5&2016\\
Video Logo Detector~\cite{liao2017mutual}&78.6&2017\\
Faster RCNN+VGG16\_D2\_M3~\cite{li2017graphic}&90.3&2017\\
Faster RCNN+CAL~\cite{8su2018open}&74.9&2018\\
% Faster RCNN\_Mobilenet~\cite{mudumbi2019approach}&92.4&2019\\
Deep Saliency Map~\cite{2020Logo}&75.8&2020\\
Logo-Yolo~\cite{4wang2020}&76.1&2020\\
LogoNet~\cite{jain2021logonet}&82.2&2021\\
OSF-Logo~\cite{meng2021adaptive}&87.0&2021\\
RetinaNet~\cite{sahel2021logo}&40.0&2021\\
RCNN~\cite{sahel2021logo}&65.0&2021\\
Faster RCNN~\cite{sahel2021logo}&74.0&2021\\
Scaled YOLOv4~\cite{palevcek2021logo}&80.4&2021\\
MFDNet~\cite{Hou2021FoodLogoDet1500AD}&86.2&2021\\

\toprule[1pt]
\end{tabular}}
\end{math}
\end{center}
\end{table}

\begin{table}[!t]
\begin{center}
\caption{Performance comparison of logo detectors on the QMUL-OpenLogo benchmark.}
\begin{math}
% \label{Logo_dataset_compare}
% \centering
\setlength{\tabcolsep}{5.5mm}{
\scriptsize
\begin{tabular}{ccc}
\toprule[1pt]
\textbf{Method}&\textbf{mAP(\%)}&\textbf{Year} \\ 
\toprule[1pt]
YOLOv2+CAL~\cite{8su2018open}&49.2&2018\\
Faster RCNN+CAL~\cite{8su2018open}&51.0&2018\\
Logo-Yolo~\cite{4wang2020}&53.2&2020\\
Faster RCNN+CAL+MPCC~\cite{su2021multi}&49.4&2021\\
OSF-Logo~\cite{meng2021adaptive}&53.3&2021\\
Scaled YOLOv4~\cite{palevcek2021logo}&61.9&2021\\
MFDNet~\cite{Hou2021FoodLogoDet1500AD}&51.3&2021\\
\toprule[1pt]
\end{tabular}}
\end{math}
\end{center}
\end{table}

\begin{table}[!t]
\begin{center}
\caption{Performance comparison of logo detectors on the Open Brands benchmark.}
\begin{math}
% \label{Logo_dataset_compare}
% \centering
\setlength{\tabcolsep}{5.8mm}{
\scriptsize
\begin{tabular}{ccc}
\toprule[1pt]
\textbf{Method}&\textbf{mAP(\%)}&\textbf{Year} \\ 
\toprule[1pt]
Brand Net (SMA)~\cite{7jin2020open}&60.4 &2020\\
Faster RCNN+RFS+MST~\cite{chen2021robust}&64.6&2021\\
Cascade RCNN+Soft NMS~\cite{jia2021effective}&65.1&2021\\
Green Hand~\cite{xu2021simple}&65.5&2021\\
Cascade RCNN+DCN v2~\cite{leng2021gradient}&70.2&2021\\

\toprule[1pt]
\end{tabular}}
\end{math}
\end{center}
\end{table}
\subsection{Intelligent Transportation}
In intelligent transportation, vehicle logo detection is significantly growing as it can effectively assist vehicle management. In addition, due to the increasing number of vehicle violations and traffic accidents, the realization of intelligent transportation systems is crucial. However, in the real world, the detection of vehicle logos is seriously disturbed due to occlusion, illumination, and low image resolution. Therefore, an efficient vehicle logo detection system is essential. In recent years, methods for vehicle logo detection have been proposed successively~\cite{22pan2013,17Zhang2013,25sotheeswaran2014,23xiang2016,15gopinathan2018,19liu2019large,yang2019fast,vehicle}. Lu \emph{et al}.~\cite{lu2021category} designed a feature extraction module VLF-Net and category-consistent mask learning module for vehicle logo detection.
Yu \emph{et al}.~\cite{26Yu2021} proposed a cascaded deep convolutional neural network to detect vehicles without relying on the presence of license plates. The network is a two-stage framework including two components: a region proposal network and a convolutional capsule network. The former is responsible for generating region proposals that may contain vehicle logos, while the latter is responsible for classifying the proposals into background and vehicle logos.  
Surwase \emph{et al}.~\cite{surwase2021multi} proposed a multi-scale multi-stream deep network for vehicle logo detection. The network processes the input image through a multi-scale stream to extract robust features, followed by logo recognition. The network follows a knowledge sharing strategy, where the learned features are shared on each stream of the network.
 \begin{figure*}[htbp]
		\centerline {\includegraphics[width=18cm]{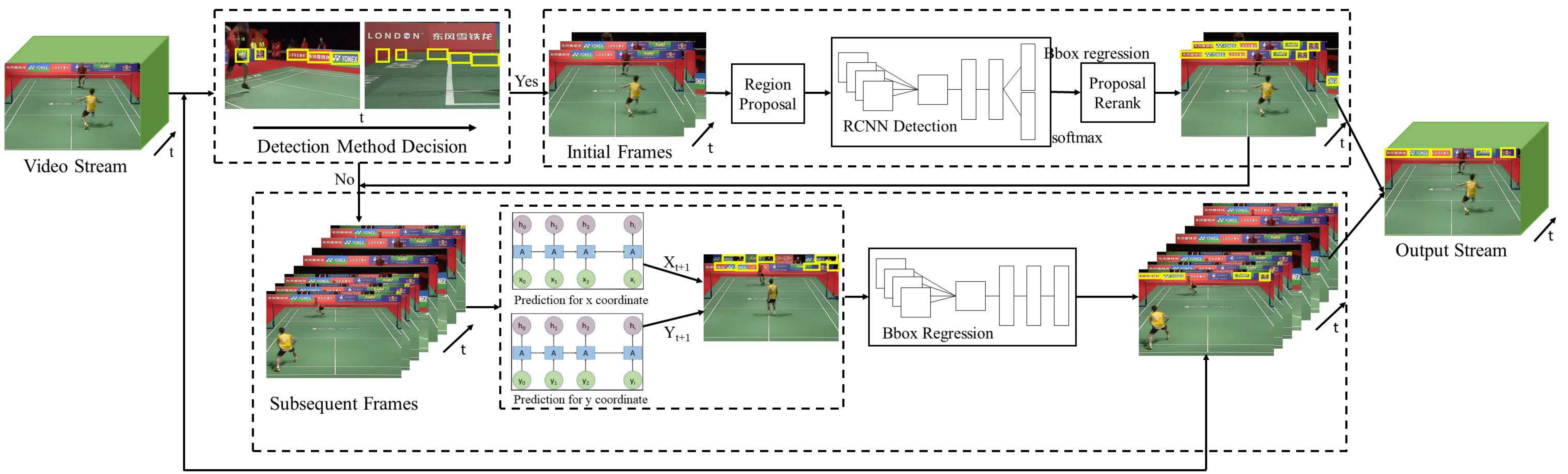}}
		\caption{The architecture for video logo detection~\cite{liao2017mutual}.}	
		\label{figLabel}
\end{figure*}

\subsection{Copyright and Trademark Compliance} 
With the development of global e-commerce platforms, the logos of major brands have become an important element in the e-commerce market. Some attackers use illegal means to infringe on the original author, so protecting intellectual property rights has also become a focus. Logo detection plays a vital role in preventing the increasing number of counterfeit transaction attempts online~\cite{7Lei2012,6Zhang2014,3guru2017symbolic,chen2021robust,jia2021effective,leng2021gradient,xu2021simple}. Infringement detection of logos has been studied for images and videos.
Chen \emph{et al}.~\cite{chen2021robust} built a highly optimized and robust detector by using techniques such as data augmentation for the detection of logos (515 categories) in e-commerce images.
Patalappa \emph{et al}.~\cite{patalappa2021robust} extended the dataset through data augmentation techniques. They applied the SSD algorithm to detect broadcast logos in the context of broadcast and broadband content piracy and proved the effectiveness of the application of SSD in content piracy video analysis monitoring tasks.
An efficient way to identify the source of a web video is to detect some specified logos. In order to identify the source of illegal videos, Ye \emph{et al}.~\cite{18Ye2017} proposed a logo detection system “GeLoGo” to detect logos in Web-scale videos. The system consists of four main modules: the keyframe module, the proposal module, the spatial verification module, and the temporal verification module. The keyframe module first extracts several frames with short fragment detection, which can significantly reduce the data. Secondly, the proposal module extracts candidate boxes through a set of pre-trained identity detectors. Then, the spatial validation module detects logos through the ResNet network. Finally, the temporal verification module detects the detection result by detecting the recognition position.

\subsection{Document Categorization}
A logo often appears in commercial documents and can be used as a statement of ownership of the document. Automatic logo detection is growing due to the increasing requirements of intelligent document image analysis and retrieval~\cite{document3,document4,document5}. 
Alaei \emph{et al}.~\cite{document1} proposed a complete system for logo detection in document images. They proposed a template-based recognition strategy for under or over-segmentation problems that may arise during detection. In order to speed up the matching of templates in the recognition phase, a search space reduction strategy that utilizes the geometric properties of logo-patches and template logo-models is proposed to reduce the number of template logo-models.  
Zhu \emph{et al}.~\cite{document2} proposed a multi-scale method for document image logo detection and extraction. The authors used an augmentation strategy across multiple image scales to classify and localize logos accurately.

\subsection{Other Applications}
Logo detection can offer various benefits for advertisers looking to improve their marketing strategies effectively. As shown in Fig. 7, Liao \emph{et al}.~\cite{liao2017mutual} proposed a video logo detection framework that used a mutual-enhanced approach to improve logo detection through the information obtained from other simultaneously occurred logos. Based on the Fast R-CNN model, a uniformity-enhanced reordering method is proposed to improve the accuracy of in-frame region suggestions by analyzing the features of logos in videos. In addition, they proposed a frame propagation enhancement method to assist in the detection of adjacent frames. Finally, the effectiveness of the method is demonstrated on the FlickrLogos-32 dataset and the video dataset.
Recently, food logo detection~\cite{Hou2021FoodLogoDet1500AD} has become more popular because it has many valuable applications, such as food brand management and displaying nutritional information.
Logo detection is also used in other areas, such as video semantic annotation~\cite{10Nedret2009,21zhang2017tv,palevcek2021logo}, and noise recognition~\cite{13wang2018,sathiaseelan2021logo,8Chen2011,14pimkote2018}. 
Since the TV logo photographed is susceptible to illumination, occlusion, noise, and other factors, the TV logo may only take up a small part of each image. To this end, Pan \emph{et al}.~\cite{4pan2016tv} proposed an efficient CNN to solve this problem. They used the maximum stable extremal region (MSER) algorithm to extract candidate frames. Since this algorithm tends to produce a large number of candidate frames without TV logos, they designed certain geometric constraints to remove non-logo objects. Finally, the images were fed into the CNN network for classification and achieved good results.

\begin{figure}[!t]  
	\centering
		\subfigure[Small size.]{
		\includegraphics[width=8.6cm]{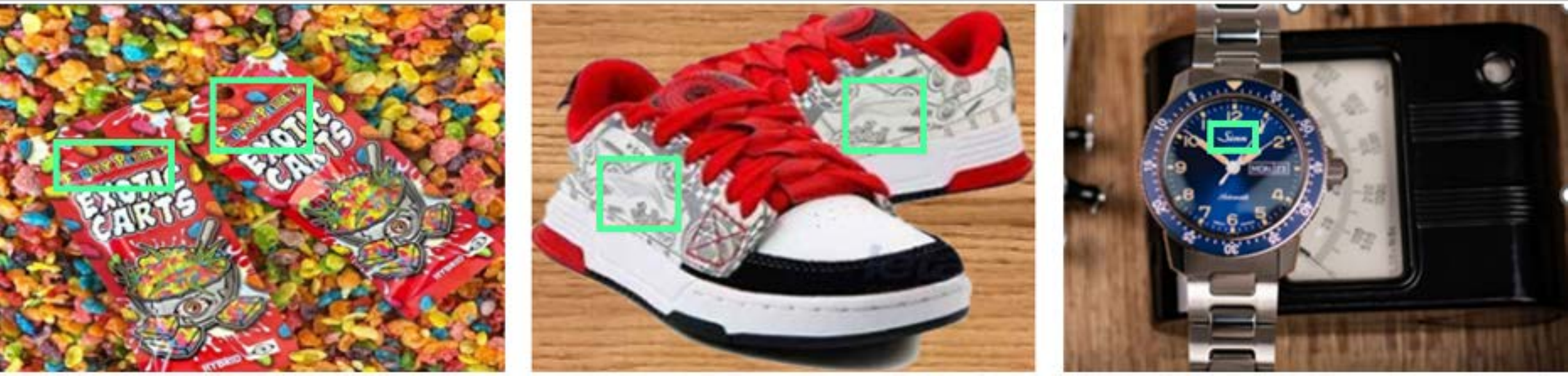}
	}
	\subfigure[Highly diverse backgrounds.]{
		\includegraphics[width=8.6cm]{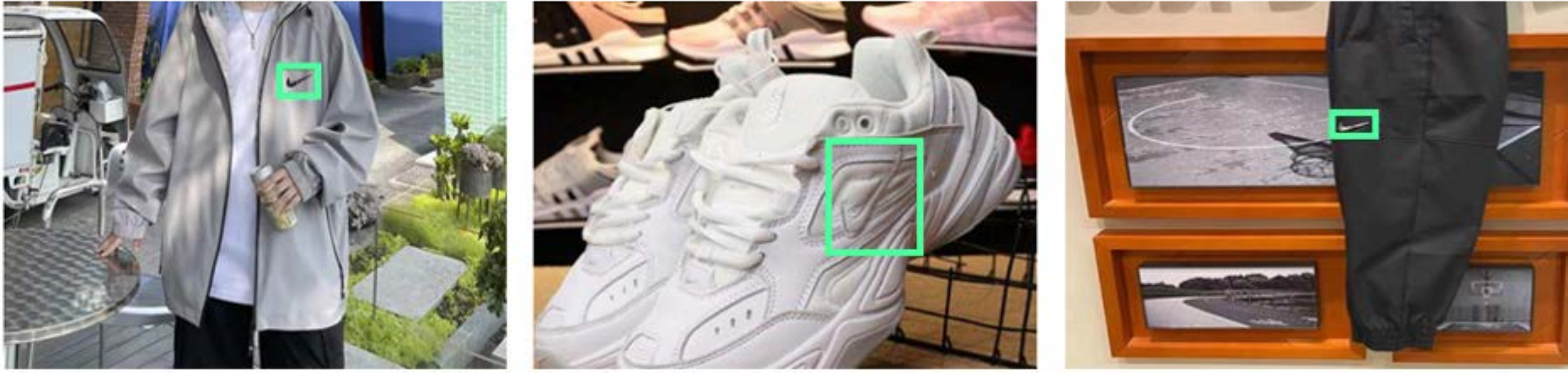}
		%\caption{fig1}
	}	
	%	\quad
	\subfigure[Sub-branding.]{
		\includegraphics[width=8.6cm]{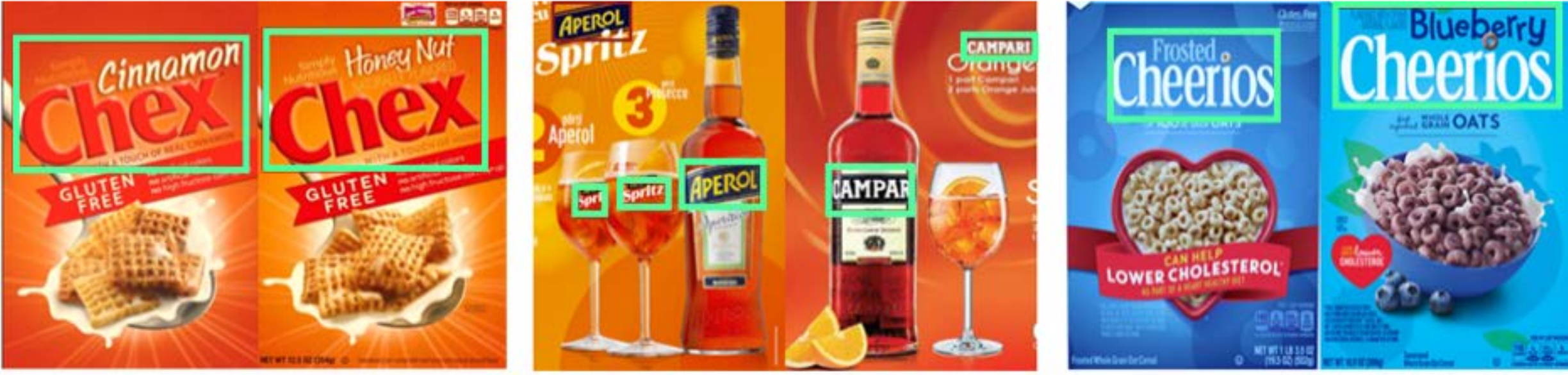}
	}	
	
	\caption{Distinctive properties of logos.}
\end{figure}

\section{Challenges and Future Directions}

\subsection{Challenges}
Logo detection has received more attention in the last few years for its wide applications. 
However, robust and accurate logo detection is still challenging in real-world scenarios because the logo images have distinctive properties, which may form significant obstacles to current progress. We try to summarize them as following: 
(1) \emph{Small size:} Unlike generic objects, logos tend to be small in size, making it difficult to distinguish them from contexts, especially the complex background. Although contextual information is crucial in small object detection because small objects carry limited information~\cite{Divvala2009AnES}, it is not always beneficial for logo detection, especially when it is too complex and unrelated to the target logo, as shown in Fig. 8(a). We know that in the deep learning-based detection architecture, different layers of feature maps with various spatial resolutions are produced due to pooling and subsampling processes. Early-layer feature maps represent small reception fields but lack high-level semantic information critical for logo detection. In contrast, high-level feature maps help identify large logos but may not detect smaller ones.
(2) \emph{Highly diverse backgrounds:} A logo is usually associated with other visible entities, like a particular service or product. Thus, a logo may appear in various situations superimposed on objects such as geometrical renderings, shirts of persons, jerseys of players, boards of shops, billboards, or posters on sports playfields. As shown in Fig. 8 (b), a brand like “Nike” may serve many types of products, such as jackets, shoes, or trousers, resulting in the same logo being attached to a variety of different objects, which poses challenges to designing a robust logo detector when considering image statistics from the entire image.
(3) \emph{Sub-branding:} Sub-branding is often considered when a company releases a new product. Sub-brand detection facilitates brand monitoring but suffers from the same problems as fine-grained classification, such as subtle differences between sub-brands and parent brands and between sub-brands. In real logo detection, some sub-brands are treated as logos different from the parent brand or other sub-brands because they have customer expectations and personalities that are different from the parent brand. Thus, the parent brand and its sub-brands usually have remarkably similar context information (background, packaging, etc., as shown in Fig. 8(c)). When considering image-level feature maps, in this case, there are probably several remarkably similar feature maps extracted from different logo images, which challenges identifying different sub-brand logos accurately when using the deep learning-based detector. 

Many approaches to tackle logo detection tasks have been developed based on object detection methods and have achieved promising results, but there is still much room for improvement. For example, \cite{yang2019fast} proposed a vehicle logo detection system based on YOLOv3. Though the performance of small logo detection has been improved to a certain extent, it fails to accurately detect small logos with complex letter patterns. \cite{new1}  designed an improved anchor-based scheme for small logos by utilizing higher resolution feature maps. However, it is too computationally expensive for constrained scenarios and too slow for real-time applications due to adopting a two-phase detection strategy.
%In addition, the generalization and robustness of these models are also challenging since the data quality significantly influences the generalization and robustness of models. The current dataset annotation work is mainly completed manually, which is easily disturbed by human subjective factors to a large extent. Thus, it is still challenging to obtain large-scale datasets with high-quality annotation.  
It is worth mentioning that the best detection performance only reaches 61.9\% of mAP on QMUL-OpenLogo~\cite{palevcek2021logo}, in which the logo detector is designed based on YOLOv4. Therefore, simply borrowing the existing detectors from general object detection is considered insufficient for logo detection. Designing a new detection paradigm that is suitable for logos is essential.

\subsection{Future Directions}
Some future research directions in logo detection are as follows:

\textbf{Lightweight logo detection:} Lightweight networks are designed to reduce model complexity further while maintaining model accuracy. With the rapid development of mobile terminals such as mobile phones and computers, people pay more attention to lightweight detectors. Lightweight logo detection can help consumers quickly identify and accurately search for the products they need and thus can significantly improve the efficiency of online shopping. Although significant efforts have been made recently, the speed gap between the machine and the human eye is still huge. The detection accuracy is also insufficient, especially for small logos. Therefore, improving the accuracy of lightweight logo detectors is one of the future development trends.

\textbf{Weakly supervised logo detection:} A fully supervised logo detector must be trained on fully labeled large-scale logo datasets. However, the labeling of existing datasets is mainly done manually, which poses a challenge to the generalization and robustness of existing models due to the problem of data quality. With the rapid growth of logo data volume, data annotation costs are getting higher and higher. Developing weakly supervised logo detection techniques that only use image-level annotations or partial bounding box annotations to train logo detectors significantly reduces costs and improves detection flexibility. Therefore, developing weakly supervised logo detection methods without fully labeled training data is an important issue for future research.

\textbf{Video logo detection:} Logo promotion exists in most video advertisements. Accurate and reasonable advertising is targets businesses, so it is essential to identify the target logo effectively. Current logo detectors are commonly used to detect logos of individual images, which may lead to a lack of correlation between consecutive images. In addition, there are problems such as motion blur and target occlusion in logo detection of videos, which can highly affect the performance of the detector. There are already methods trying to detect video logos. Currently, there are methods~\cite{Su2022FewST,liao2017mutual} trying to detect video logos, but they have not achieved satisfactory results. Therefore, it is an important research direction to improve logo detection by utilizing temporal and spatial correlations.

\textbf{Tiny logo detection:} Tiny logos contain insufficient information, making it more challenging to extract discriminative features. Meanwhile, tiny logos contain relatively few samples and thus are easily disturbed by environmental factors. Numerous real-world tiny logos may have highly diverse contexts, which can cause the same logo to appear very different in different real scenarios. Observing the results of existing detection methods on tiny logos, we can find that the performance of this methods~\cite{velazquez2021logo,yang2019fast,zhang2021vehicle} is unsatisfactory in real scenarios.

\textbf{Long tail logo detection:} With the rapid development of society, some enterprises are currently leading the way, and their enterprise logos naturally often appear in front of the public. In contrast, the logos of most small enterprises are not paid much attention to because they appear less. The long-tail distribution of logos exists all the time in reality. Some logo classes contain a large number of samples, while others contain few samples, making it difficult for logo detectors to detect logo classes with few samples. In previous logo detection methods, there is little focus on the long-tailed distribution in logo images. Therefore, it is one of the future tasks to utilize the unbalanced data to train efficient and accurate detectors.

\textbf{Incremental logo detection:} As a company's signature or symbol, the logo will continue to increase as the number of companies increases. Therefore, the logo detector must learn new logos. However, retraining the logo detector is time-consuming and resource-intensive by discarding the previously learned knowledge. Incremental learning enables a system to continuously learn new knowledge from new samples while retaining most previously learned knowledge without catastrophic forgetting. It can reduce the complexity of the model to a certain extent, thereby reducing the training cost of the model. However, no incremental learning method currently can show good performance under all conditions. Therefore, incremental learning for logo detection is one of the future research directions.

In fact, existing studies generally assume close-environment scenarios. For example, the data distribution of logo datasets is invariable, and all the training data have known classes and fine-grained bounding box annotations in advance. Although Su \emph{et al}.~\cite{8su2018open} proposed an assumption to simulate the open deployment, they mainly focus on the problem of limited logo data by presenting a context adversarial learning approach to generate context-consistent synthetic training data automatically. How to enable the trained model to be updated according to emerging new logo classes or data distribution change, or even we hardly know what changes is an essential requirement in open-environment.

\section{Conclusion}
In this work, we offer a comprehensive survey of previous and current approaches that helps logo detection. We mainly review the recent advances in deep learning-based logo detection by summarizing classical solutions. In addition, we comprehensively review the commonly used datasets, summarize the related applications of logo detection, and predict future research directions. Although the achievement of logo detection has been effective recently, there is still much room for further development. We hope this paper will better inform readers about the current development of logo detection and inspire more people to get involved in logo detection.

\footnotesize
\bibliographystyle{IEEEtran}
\bibliography{Logo_detection}  % sigproc.bib is the name of the Bibliography in this case

\end{document}